\documentclass[preprint,12pt]{elsarticle}

\usepackage{wrapfig}
\usepackage{booktabs}
\usepackage{caption}
\usepackage{subcaption}
\usepackage{float}% If comment this, figure moves to Page 2
\usepackage{multirow}
\usepackage{hyperref}

\graphicspath{{./img/}}

\usepackage{amsmath}
\usepackage{amssymb}
\usepackage{lineno}

\journal{Smart Agricultural Technology}

\begin{document}

\begin{frontmatter}

%% Title, authors and addresses
\title{Enhancing annotations for 5D apple pose estimation through 3D Gaussian Splatting (3DGS)}

\author[inst1]{Robert van de Ven}
\ead{robert.vandeven@wur.nl}
\author[inst2]{Trim Bresilla}
\author[inst1]{Bram Nelissen}
\author[inst2]{Ard Nieuwenhuizen}
\author[inst1]{Eldert J. van Henten}
\author[inst1]{Gert Kootstra}

\affiliation[inst1]{organization={Agricultural Biosystems Engineering, Wageningen University \& Research},
            addressline={Droevendaalsesteeg 1}, 
            city={Wageningen},
            postcode={6708 PB},
            country={the Netherlands}}
\affiliation[inst2]{organization={Agrosystems Research, Wageningen University \& Research},
            addressline={Droevendaalsesteeg 1}, 
            city={Wageningen},
            postcode={6708 PB},
            country={the Netherlands}}

\begin{abstract}
%% Text of abstract
Automating tasks in orchards is challenging because of the large amount of variation in the environment and occlusions. One of the challenges is apple pose estimation, where key points, such as the calyx, are often occluded. Recently developed pose estimation methods no longer rely on these key points, but still require them for annotations, making annotating challenging and time-consuming.
Due to the abovementioned occlusions, there can be conflicting and missing annotations of the same fruit between different images. Novel 3D reconstruction methods can be used to simplify annotating and enlarge datasets. 
We propose a novel pipeline consisting of 3D Gaussian Splatting to reconstruct an orchard scene, simplified annotations, automated projection of the annotations to images, and the training and evaluation of a pose estimation method. 
Using our pipeline, 105 manual annotations were required to obtain 28,191 training labels, a reduction of 99.6\%. Experimental results indicated that training with labels of fruits that are \(\leq95\%\) occluded resulted in the best performance, with a neutral F1 score of 0.927 on the original images and 0.970 on the rendered images. 
Adjusting the size of the training dataset had small effects on the model performance in terms of F1 score and pose estimation accuracy.  It was found that the least occluded fruits had the best position estimation, which worsened as the fruits became more occluded. It was also found that the tested pose estimation method was unable to correctly learn the orientation estimation of apples.
\end{abstract}

%%Research highlights
\begin{highlights}
\item Novel pipeline simplifying pose annotation
\item Novel method to quantify the occlusion rate was developed
\item 99.6\% reduction in the amount of manual annotations
\item Training with an occlusion rate \(\leq95\%\) for the labels lead to the best performance
\item Improved fruit detection and similar pose estimation as state of the art
\end{highlights}

\begin{keyword}
%% keywords here, in the form: keyword \sep keyword
Fruit pose \sep Gaussian Splatting \sep 3D detection \sep Fruit occlusions
%% PACS codes here, in the form: \PACS code \sep code
% \PACS 0000 \sep 1111
%% MSC codes here, in the form: \MSC code \sep code
%% or \MSC[2008] code \sep code (2000 is the default)
% \MSC 0000 \sep 1111
\end{keyword}

\end{frontmatter}

% \linenumbers

%% main text
\section{Introduction}\label{sec:introduction}
Due to reduced labor availability and increased labor costs, there has been an increasing interest in automation in orchards through the use of robotics \cite{RN103}. This is in particular the case for harvesting, as it requires a large amount of manual labor in a short period of time. Automating tasks in orchards, however, is challenging because of the large amount of variation in the environment and the frequent occurrence of occlusions \cite{RN153}. For the harvesting of apples, a specific challenge is the correct detachment of the fruits, which requires a specific rotational motion for optimal detachment at the abscission layer \cite{RN64,RN164,RN190}. This requires information about the position and orientation, the pose, of the fruit. 

Due to the rotational symmetry around the center axis, the pose of an apple, as for many other fruits, can be defined as 5-dimensional (5D), consisting of the 3D position and the 2D orientation of the center axis.  
Previous work on apple-pose estimation used multi-stage \cite{RN205,RN252} or end-to-end detectors \cite{RN226}. Multi-stage detectors separate fruit detection and pose estimation, often utilizing separate neural networks for the detection of the fruit and the detection of key points for the calyx or peduncle, based on which a fruit pose is calculated. With multi-stage detectors, the fruits can be detected quite well, while the smaller calyx or peduncle is much more challenging to detect. \citet{RN252} achieved an Average Precision (AP) between IoU thresholds of 0.5 to 0.95 (AP@0.5:0.95) of 0.944 for the apple, but only 0.583 for the calyx. In a lab environment, it was found that the average orientation error was 12.3°, with increasing error when the fruits were less visible. In a real orchard, \citet{RN205} achieved a median orientation error of 17.6°. 
If neither the calyx nor peduncle is visible or detected, these multi-stage detectors cannot estimate the orientation, which is the case in a substantial amount of situations. 
In comparison, end-to-end detectors use a single network to perform both the detection and pose estimation in a single pass. These detectors do not rely on the detection of specific key points to determine the orientation of the fruit, allowing them, in principle, to estimate the orientation of fruits without the key points visible. \citet{RN226} developed a deep neural network, {FRESHNet}, which predicted 3D oriented bounding boxes per apple instance in RGB-D images, including a custom stem direction loss, which accounted for the rotational symmetry to improve orientation estimation. They used a dataset with 312 unique apples captured in 3,381 images, resulting in 9,044 apple instances in the images. FRESHNet achieved an mAP@0.5 of 0.783, and a pitch and yaw error of 23.9°, and 34.4° respectively. This method was not yet able to improve the performance compared to the multi-stage approaches. A potential reason is the limited size of the dataset in comparison to that used in \citet{RN269}, on which {FRESHNet} was based. 

High quality training data is essential for the correct functioning of the end-to-end method. While it is straightforward to annotate instance masks for fruit detection, it is much more challenging to accurately annotate the pose, specifically the orientation, of fruits in a camera image. Especially when neither the peduncle nor calyx is visible, the orientation cannot reliably be annotated. In addition, as in \citet{RN226}, datasets typically consist of a large set of images taken from a smaller set of fruits under different viewpoints. Annotating all instances in the images is very time-consuming, and the consistency of labeling the orientation of the same apples in multiple images is typically low, resulting in conflicting annotations, which limit the performance of the trained model. Furthermore, there is an increased risk of data leakage, as the same apple occurs in multiple images, which should not end up in the training and the test set. 

In this paper, we propose the use of 3D scene reconstruction from camera images to deal with the above-mentioned challenges in data collection, data annotation, and model training. In the 3D reconstruction, the data is more complete through the integration of multiple viewpoints. This allows better labeling of the fruit pose, as the annotator can rotate the 3D scene in the annotation process. A 3D scene reconstruction, furthermore, allows the rendering of new camera views.

However, the quality of the rendered images relies on the quality of the reconstruction. Traditional methods, like Structure from Motion (SfM) and multi-view stereo (MVS) lack density to achieve a high quality reconstruction \cite{RN268}. As a result, the technique was used to render depth for existing images instead of rendering additional images, such as in \citet{RN265,RN266}. However, recent deep learning methods resulted in a major leap forward in the quality of the reconstruction \cite{RN238}. 
Especially 3D Gaussian Splatting (3DGS) is promising, with reduced computational intensity for rendering, enabling real-time rendering of high quality images \cite{RN238}. 
These novel 3D scene reconstruction methods can be used to address challenges in fruit pose estimation by viewing around occlusions to improve annotations and render additional images from novel viewpoints to enlarge and enrich the dataset. 

In this paper, we propose a novel pipeline, consisting of 3DGS to reconstruct an orchard scene, simplified annotations, automated projection of the annotations to images, and the training and evaluation of a pose-estimation method. 
The annotation of the apple poses is done using the 3DGS. These annotations are then projected onto the images to automatically create annotated image data. This greatly simplifies the process of obtaining 3D annotations, as only the unique apples need to be labeled instead of all apples in all images. Moreover, labels are now available for all fruits regardless of possible occlusions in a given image, while a human annotator would be unable to annotate a fruit in a 2D image if the occlusions are above a certain level. In addition, the 3DGS enables rendering additional training data from novel viewpoints to enlarge and enrich the dataset. The generated image data is then used to train and test an apple detection and pose estimation method. Through experiments, we investigated the effect of occlusion rate on the model performance, and the effect of dataset size -- including rendered data -- on model performance. 

\section{Materials \& Methods}\label{sec:mm}
In this section, we describe the proposed pipeline, the dataset, and the experiments. 

\subsection{Proposed pipeline}
\begin{figure}
    \centering
    \includegraphics[width=\linewidth]{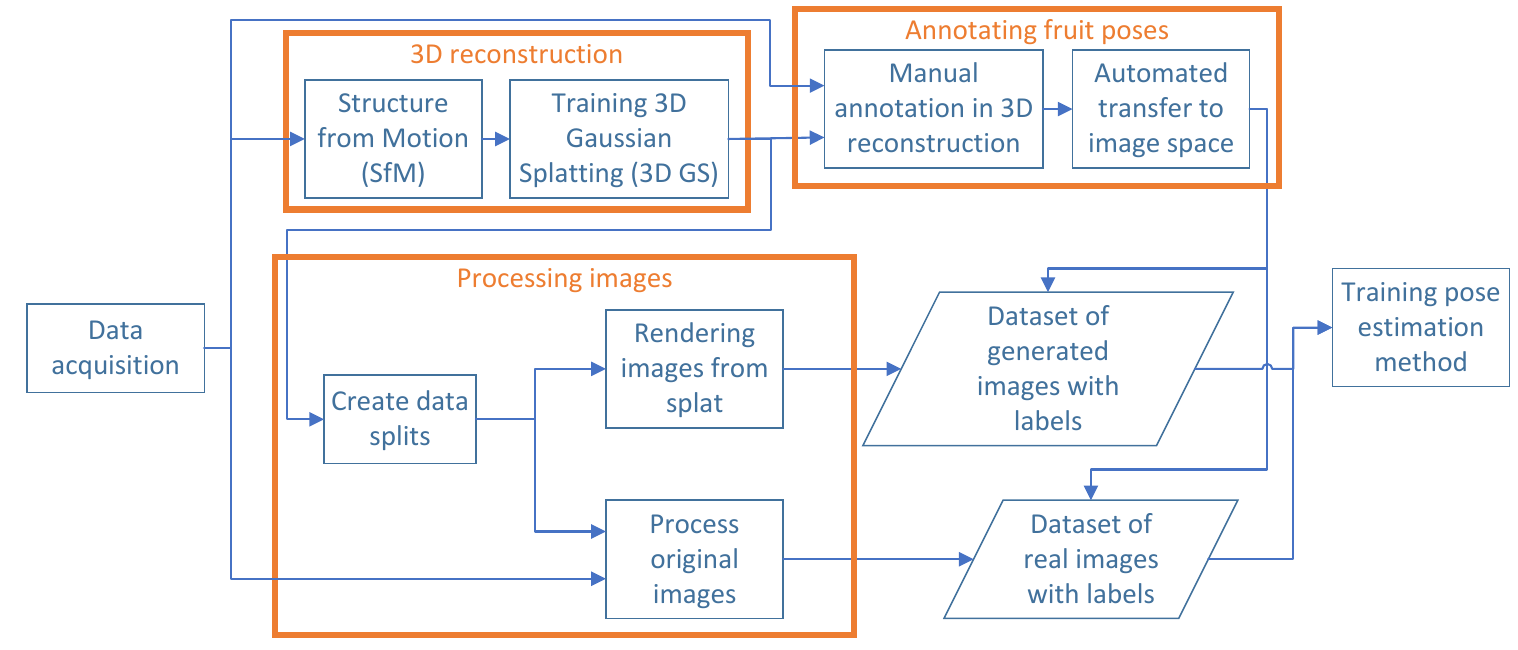}
    \caption{Flowchart showing the proposed pipeline for pose estimation. }
    \label{fig:mm:flowchart}
\end{figure}

The proposed pipeline\footnote{The pipeline is available at \url{https://github.com/WUR-ABE/Gaussian-Splatting_pose-estimation}} used in this study is shown in Figure \ref{fig:mm:flowchart}. In the pipeline, the first step is to acquire a dataset of RGB images, described in section \ref{sec:mm:data-acq}. Then, the scene is reconstructed in 3D using Structure from Motion (SfM) and 3D Gaussian Splatting (3DGS), as described in section \ref{sec:mm:3d-reconstruction}. Once the scene has been reconstructed, the fruit poses can be annotated in the reconstruction and transformed to image space, as described in section \ref{sec:mm:annotating}. Then, the data is split and images rendered or processed, as described in section \ref{sec:mm:processing}. Lastly, the created datasets can be used to train the pose estimation method, as described in section \ref{sec:mm:pose-estimation}.  

\subsubsection{Dataset acquisition}\label{sec:mm:data-acq}
The dataset collected for this paper consisted of images taken of 13 trees in an apple orchard located in Randwijk, the Netherlands. The images were taken 2 October 2024 between 16:25 and 17:05, during overcast weather conditions. The images were taken using a Nikon Z6 system camera. The exact settings are provided in \ref{sec:app:cam-settings}. To ensure that the quality of the reconstruction is as high as possible, images were taken at varying heights while moving around the set of trees, and multiple trees were included in each image. In addition, the pitch and yaw of the camera were adjusted at each height to ensure complete coverage of the trees. In total, 367 images were collected. The dataset is available at \url{https://doi.org/10.4121/976c94f2-028f-4291-adfd-20eb82b0f647}

\subsubsection{3D reconstruction through Gaussian Splatting}\label{sec:mm:3d-reconstruction}
In our study, the scene was reconstructed using 3D Gaussian Splatting (3DGS) \cite{RN238}. A well reconstructed scene enables the rendering of realistic images to train a pose estimation algorithm. In 3DGS, the scene is represented by optimized 3D Gaussians. Each 3D Gaussian has a center (position) \(\mu\), opacity \(\alpha\), 3D covariance matrix (size and orientation) \(\Sigma\), and color \(c\). The color across the surface of the Gaussian is represented using spherical harmonics to enable view-dependent appearance. 

The placement of these 3D Gaussians is determined through an optimization process utilizing back projection, which both optimizes the properties of the 3D Gaussians present and optimizes the number of 3D Gaussians present \cite{RN238}. The number of Gaussians is initialized using the sparse points from Structure from Motion (SfM), as a good initialization is essential for high reconstruction quality.  
The properties of the 3D Gaussians are optimized through back-propagation, with the covariance matrix \(\Sigma\) represented as a quaternion \(q\) and a 3D scale vector \(s\) to prevent creating a non-positive semi-definite covariance matrix. The Gaussians can be densified by splitting a large 3D Gaussian into two scaled down 3D Gaussians or copying a small 3D Gaussian and shifting the copy to a less well constructed area. In addition, the Gaussians can be pruned by removing Gaussians that are virtually transparent, i.e., with low opacity \(\alpha\), and removing Gaussians that are too large in view-space or world-space. 

\begin{figure}[t]
    \centering
    \includegraphics[width=\linewidth]{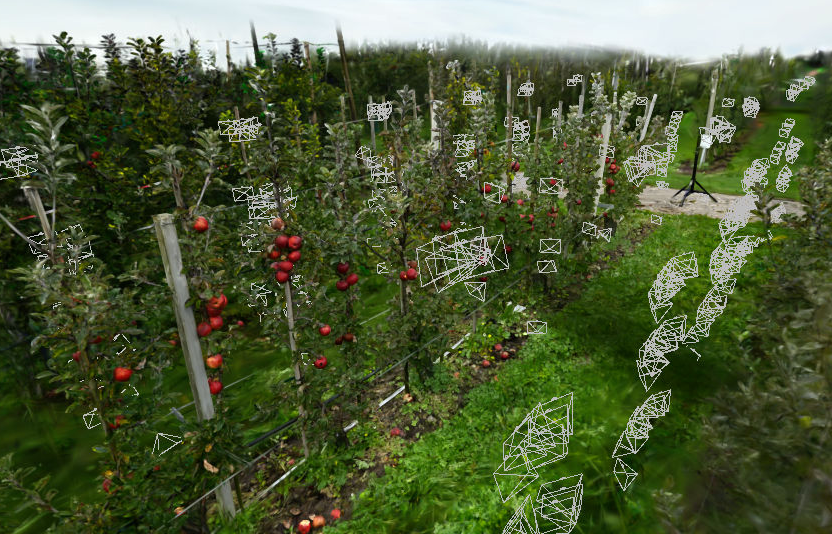}
    \caption{Image from the trained 3DGS. The camera poses are shown using frustums in white. Not all camera poses are visible from this viewpoint.}
    \label{fig:mm:trained-gs}
\end{figure}

In our case, the software Agisoft Professional Photoscan using the COLMAP algorithm \cite{RN270} was used to obtain the initial sparse point cloud. In total, 364 images were matched. The settings are shown in \ref{sec:app:sfm-settings}. Next, the camera poses were refined using the SO3xR3 optimizer from Nerfstudio \cite{RN271}. Lastly, the 3DGS was trained using the software Jawset Postshot, using the settings shown in \ref{sec:app:3dgs-settings}. The trained 3DGS is shown in Fig. \ref{fig:mm:trained-gs}. The thirteen trees that were focused on are visible in high detail, while the general structure of other trees can also be observed.

\subsubsection{Annotating fruit poses}\label{sec:mm:annotating}
The trained 3DGS, in combination with the camera poses, enables the annotation of fruit poses in the 3D reconstruction, providing a single annotation per fruit and projecting these annotations to image space for training a pose estimation method. In the following sections, we will describe the process of annotating the global fruit pose and projecting these fruit poses to image space. 

\paragraph{Manual annotation in 3D reconstruction}

\begin{figure}
    \centering
    \subfloat[3DGS]{
        \includegraphics[width=0.21\columnwidth]{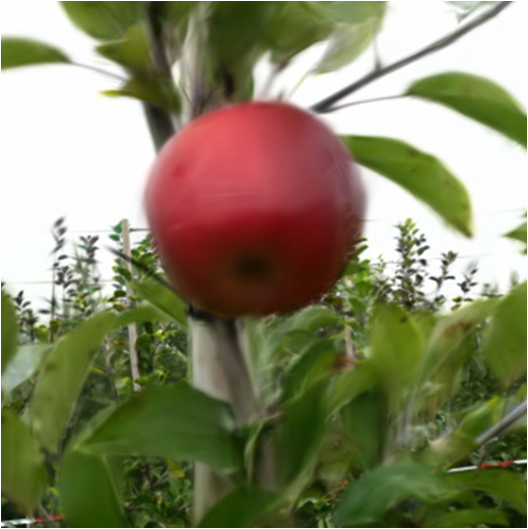}
        \label{fig:mm:global-labels:3d-gs}
    }
    \hfill
    \subfloat[Point cloud]{
        \includegraphics[width=0.21\columnwidth]{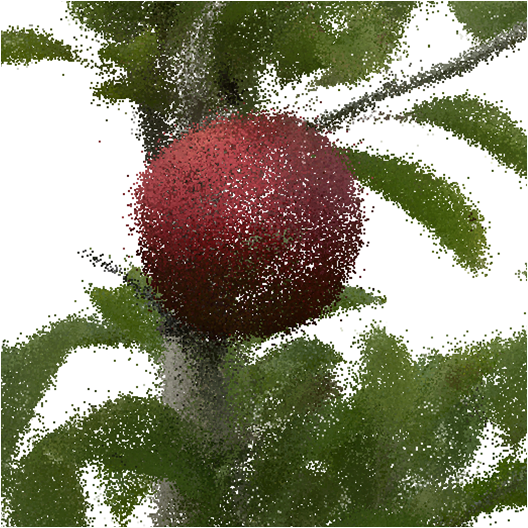}
        \label{fig:mm:global-labels:pointcloud}
    }
    \hfill
    \subfloat[Individual apple]{
        \includegraphics[width=0.21\columnwidth]{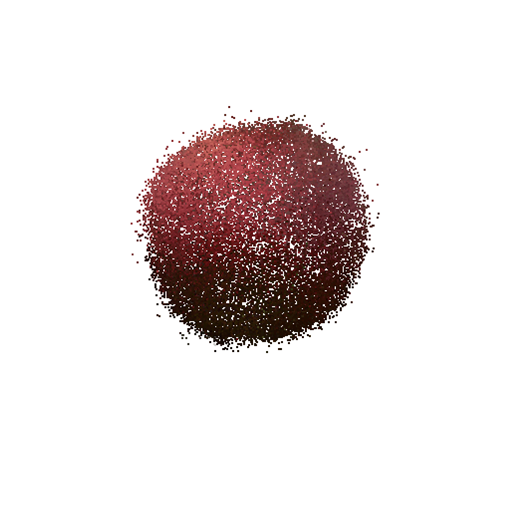}
        \label{fig:mm:global-labels:fruit}
    }
    \hfill
    \subfloat[Calyx point]{
        \includegraphics[width=0.21\columnwidth]{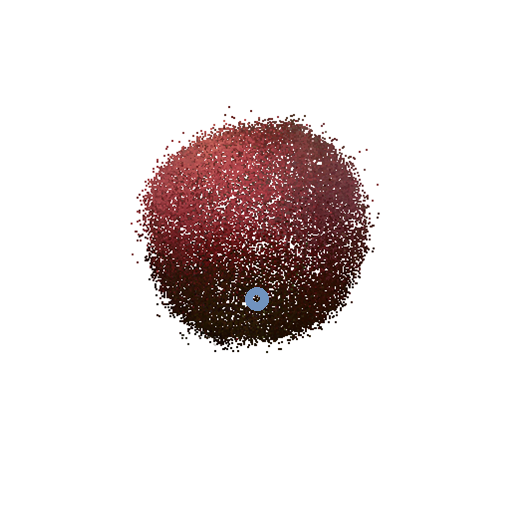}
        \label{fig:mm:global-labels:calyx}
    }
    \caption{Steps to obtain fruit pose annotations.}
    \label{fig:mm:global-labels}
\end{figure}

The steps for manual annotation are shown in Fig. \ref{fig:mm:global-labels}. Fig. \ref{fig:mm:global-labels:3d-gs} shows the 3DGS.
This 3DGS was converted to a high density point cloud using the 3DGS-to-PC toolbox \cite{RN273}, using 50,000,000 points. The resulting point cloud is shown in Fig. \ref{fig:mm:global-labels:pointcloud}.
In this point cloud, each fruit was manually segmented by cropping the point cloud from multiple viewpoints until only points belonging to the fruit remained. An example of the resulting point cloud for a single apple is shown in Fig. \ref{fig:mm:global-labels:fruit}. 
Next, the fruit orientation was determined by selecting the point in the fruit point cloud that lies at the center of the calyx. In some cases, the color in the point cloud was not clear enough to determine the calyx. In these cases, the trained 3DGS and the original images were used to determine where the calyx was located. The selected point of the calyx location is circled in blue in Fig. \ref{fig:mm:global-labels:calyx}. 

\paragraph{Automated transfer to image space}
As the camera intrinsics and extrinsics are known, the fruit point clouds can be transformed into each image. For the fruits that are present in an image, the required label for the pose estimation method is calculated. In our case, this consists of a 2D bounding box in the image and a 3D oriented bounding box, using the calyx to determine the orientation. In addition, the calyx key point or a mask could also be created. 

\begin{figure}
    \centering
    \subfloat[98\%]{
        \includegraphics[width=0.12\columnwidth]{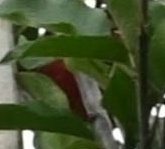}
        \label{fig:example-98}
    }
    \hfill
    \subfloat[80\%]{
        \includegraphics[width=0.12\columnwidth]{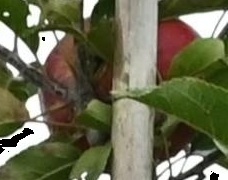}
        \label{fig:example-80}
    }
    \hfill
    \subfloat[60\%]{
        \includegraphics[width=0.12\columnwidth]{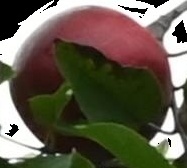}
        \label{fig:example-60}
    }
    \hfill
    \subfloat[40\%]{
        \includegraphics[width=0.12\columnwidth]{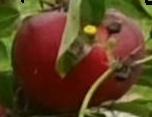}
        \label{fig:example-40}
    }
    \hfill
    \subfloat[20\%]{
        \includegraphics[width=0.12\columnwidth]{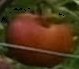}
        \label{fig:example-20}
    }
    \hfill
    \subfloat[16\%]{
        \includegraphics[width=0.12\columnwidth]{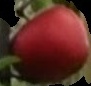}
        \label{fig:example-16}
    }
    \caption{Examples of occlusion rates of different apples}
    \label{fig:mm:examples-visibility}
\end{figure}

However, being within the camera frustum does not mean that a fruit is clearly visible, as there could be leaves in front of the fruit, or it could be on the other side of the tree. Therefore, the visibility of the fruits also needs to be determined. 
The visibility of each fruit within the frustum was determined by first rendering a depth image of the fruit point cloud using the camera intrinsics and extrinsics. As the point cloud only contains a single fruit, there are no occluding objects and all pixels with a depth belong to the fruit, resulting in a total area of \(s_T\) pixels. Next, the depth image was rendered using the 3DGS as the source. As this includes the whole scene, the occlusions are also present. Then, for each pixel that had a depth in the image from the fruit point cloud, the difference in depth with the depth image from the 3DGS was determined. If this difference was greater than 15 mm, the pixel was counted as occluded, resulting in a total occluded area of \(s_O\) pixels. The occlusion rate was then determined as: 

\begin{equation}
    o = \frac{s_O}{s_T} \times 100 [\%]
\end{equation}

\noindent where \(o\) indicates the occlusion rate of a given fruit in a given image. 
Fig. \ref{fig:mm:examples-visibility} shows fruits with different amounts of occlusion. As \(s_T\) is determined from a point cloud, there can be small areas without depth on the front of the fruit but with depth on the back of the fruit. As the difference in depth was used to determine occlusions, these areas would be falsely counted as occluded. Because of this, very few fruits had occlusion rates lower than 16\%. For example, the fruit shown in Fig. \ref{fig:example-16} does not seem to have any occlusions. 

\subsubsection{Processing images}\label{sec:mm:processing}
To create datasets from the original images and the 3DGS, several steps are required. First, the data splits need to be introduced to create clean splits and prevent data leakage. Then, novel images can be rendered using these splits and the original images can be processed to follow the data splits. The pose estimation method used in this pipeline uses RGB images and a depth image, so depth images need to be rendered as well. 

\begin{figure}[t]
    \centering
    \subfloat[Full 3DGS]{
        \includegraphics[width=0.43\columnwidth]{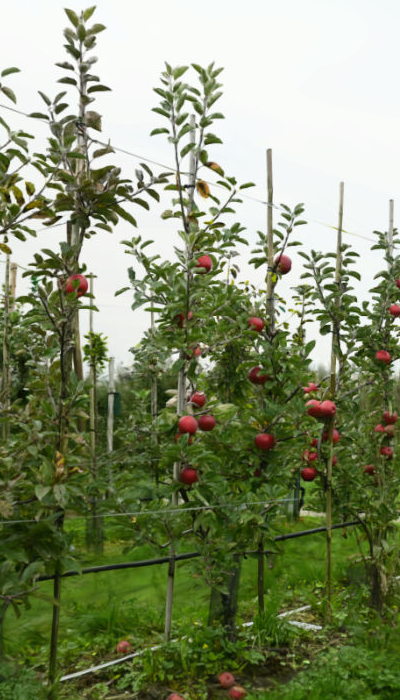}
        \label{fig:bounding-box-full}
    }
    \hfill
    \subfloat[Cropped 3DGS]{
        \includegraphics[width=0.43\columnwidth]{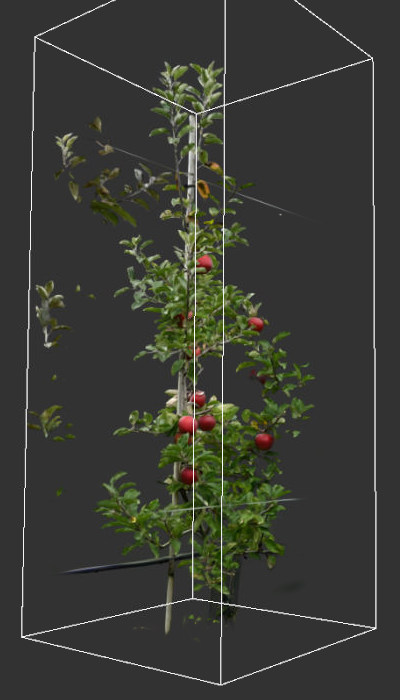}
        \label{fig:bounding-box-cropped}
    }
    \caption{Figure showing the full 3DGS and the cropped 3DGS when using the bounding box of a tree. }
    \label{fig:bounding-box}
\end{figure}

\paragraph{Creating data splits}
After the 3DGS was trained, a bounding box was drawn around each tree, separating the 3DGS into 13 smaller 3DGS, each containing only a single tree. As the trees did not exactly fall within a bounding box, the emphasis was put on ensuring the fruits of each tree were in the correct bounding box, and some leaves were allowed to fall outside the bounding box. Fig. \ref{fig:bounding-box} shows a bounding box in the 3DGS, where it can be seen that some leaves from another tree are included in the bounding box. 

\paragraph{Rendering novel images}\label{sec:mm:rendering}
Images are created by projecting all 3D Gaussians to the image space, the process called "splatting", resulting in 2D Gaussians, with an expected depth for each 2D Gaussian. Next, the Gaussians are filtered to only include the Gaussians present in the image and sorted on depth. Next, the color for each pixel is determined based on the color, depth, and opacity of each Gaussian present, where the opacity of the Gaussian with the lowest depth is used to determine the influence of the Gaussians with higher depth. 
Simultaneous with the RGB, the depth image is rendered, using the opacity normalized expected depth \cite{RN302}. 

\paragraph{Processing original images}\label{sec:mm:processing-original}
The original images included multiple fruits and lacked depth information. In order to achieve clean data splits, each fruit should only occur in a single split. This can be achieved by splitting the dataset using the trees, where the edge of the trees generally does not contain fruits, allowing clean splits without fruits on the boundary. Therefore, each image should only include a single tree. In addition, a depth image is required for the pose estimation method. Therefore, additional processing was required to achieve a usable dataset. 

\begin{figure}
    \centering
    \includegraphics[width=\linewidth]{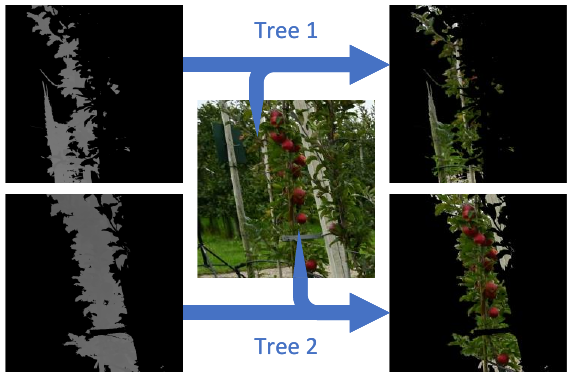}
    \caption{Masking of a single image to obtain multiple images that show only a single tree. On the left, the depth image}
    \label{fig:mm:image-masking}
\end{figure}

A depth image can be rendered in the same way as when rendering novel images, but using the intrinsics and extrinsics of the original image. However, the original images generally contain multiple trees, which can result in data leakage. In the middle of Fig. \ref{fig:mm:image-masking}, part of an original image is shown, where three different trees can be seen. To create clean data splits, the images were split into separate images for each tree present, with the pixels showing another tree masked out. 

To achieve this, two depth images were rendered. First, a depth image was rendered using the complete 3DGS, as shown in Fig. \ref{fig:bounding-box-full}. Next, a depth image was rendered using the 3DGS containing only the desired tree, as shown in Fig. \ref{fig:bounding-box-cropped}. 

The cropped depth image only shows the desired tree but lacks occlusions of other trees, which can occlude part of the leaves and fruits. Therefore, the depth in the cropped image was compared with the depth in the complete image. For any pixel where the depth in the complete image was lower than in the cropped image, the pixel in the cropped image was set to zero. This indicates that there is an occlusion and therefore the desired tree was not visible, and there could not be a depth measurement. This masked and cropped depth image shows the depth of a single tree while accounting for the occlusions caused by other trees. 

Then, the masked and cropped depth image was used to mask the original RGB image, converting the pixels without depth to black in order to hide the other trees. This process is shown in Fig. \ref{fig:mm:image-masking}. The outlines of the removed trees can be observed by the outlines of the black areas. 

\subsection{Pose estimation method}\label{sec:mm:pose-estimation}
In order to perform pose estimation, FRESHNet \cite{RN226} was used. FRESHNet uses multimodal input, where the RGB-D image is converted to a colored point cloud and a separate RGB image to enable the detection of 3D oriented bounding boxes around the fruits. FRESHNet is a multimodal detector that combines geometric information from a colored point cloud with visual features from an RGB image. Architecturally it uses parallel encoders for the point cloud and the RGB view, fuses intermediate features, and predicts object-level outputs, specifically 3D oriented bounding boxes, which contain information about location, orientation, and size. Instead of regressing a full 6-DoF orientation, the network enforces the bounding-box X-axis to align with the vector from the fruit center to the calyx; this reduces orientation prediction to a unit stem direction vector and eliminates the redundant rotation around that axis. 

Training uses a stem-direction loss that encourages the predicted unit vector \(\hat v\) to point toward the annotated calyx. This loss is combined with standard localization and size losses for the bounding box and with classification losses.

\subsection{Dataset}
The dataset acquisition was described in Section \ref{sec:mm:data-acq}. Through the proposed pipeline, we obtained two datasets, one containing the processed original images and the other containing the novel rendered images. 
For both datasets, the data was split into a train, validation and test split. The train split contained ten trees and 79 unique apples, the validation split two trees and 19 unique apples, and the test split one tree and seven unique apples.

\subsubsection{Original images dataset}
\begin{figure}
    \centering
    \includegraphics[width=\linewidth]{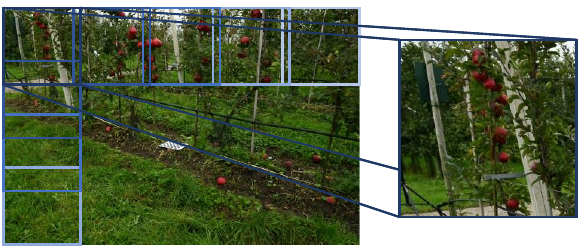}
    \caption{Patches obtained from a single image. Along the vertical axis, four rows of patches were created. Along the horizontal axis, five columns of patches were created. In the large image, the overlap between patches can be observed.}
    \label{fig:mm:image-patching}
\end{figure}

The original images were 6048 by 4024 pixels. In order to maintain high quality images for the pose estimation method, these were patched into smaller images of 1300 by 1300 pixels. As this does not scale down exactly from the original image size, each patched image overlapped slightly with the other images. Each original image was converted into 20 smaller patched images. Fig. \ref{fig:mm:image-patching} shows an example of the patching of images, showing the horizontal and vertical overlap between the patched images. In order to maintain correct camera intrinsics of the patched images, the center point was shifted according to the patch location. 

\begin{figure}[t]
    \centering
    \subfloat[Train orientations]{
        \includegraphics[width=0.3\columnwidth]{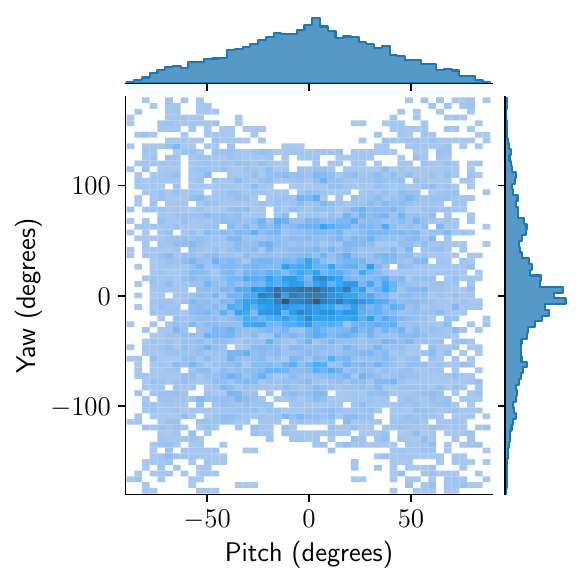}
        \label{fig:original-orientation-train}
    }
    \hfill
    \subfloat[Validation orientations]{
        \includegraphics[width=0.3\columnwidth]{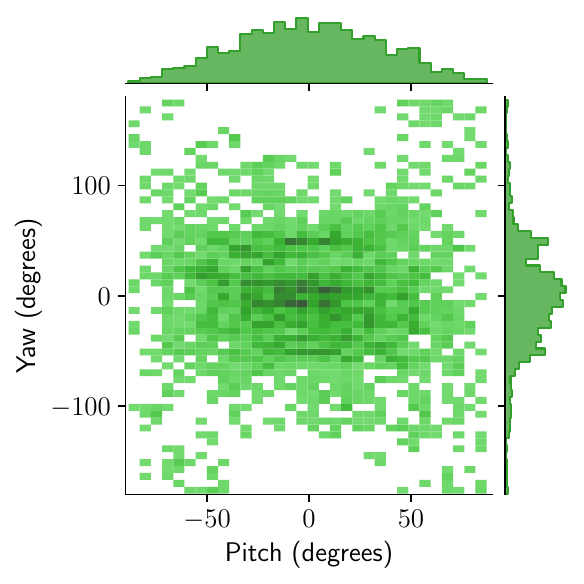}
        \label{fig:original-orientation-val}
    }
    \hfill
    \subfloat[Test orientations]{
        \includegraphics[width=0.3\columnwidth]{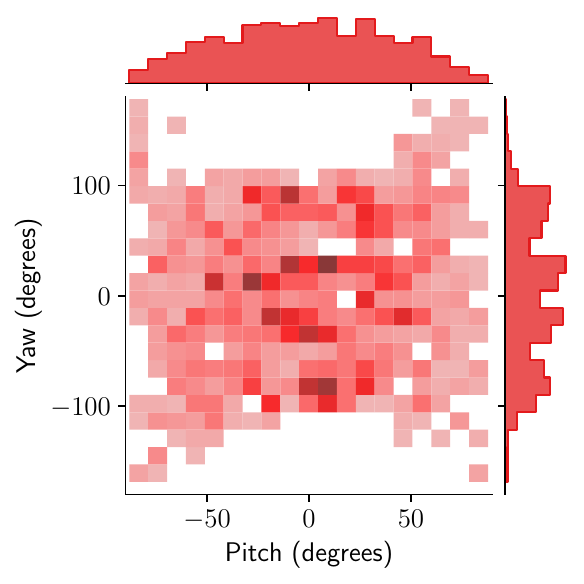}
        \label{fig:original-orientation-test}
    }
    \hfill
    \subfloat[3D view of orientations]{
        \includegraphics[width=0.45\columnwidth]{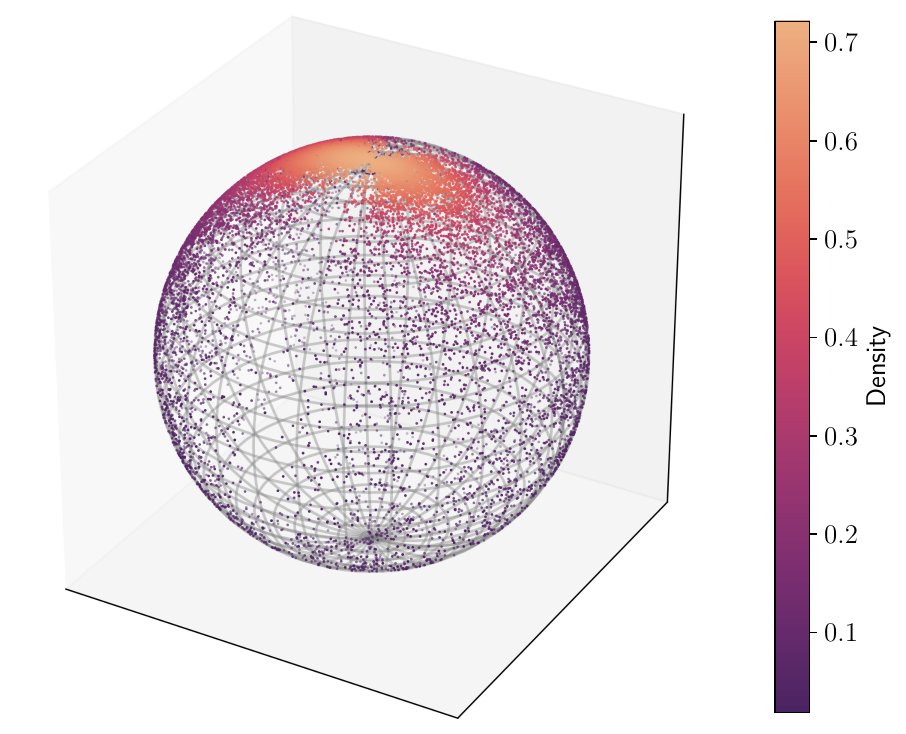}
        \label{fig:original-orientation-3d}
    }
    \hfill
    \subfloat[Occlusions]{
        \includegraphics[width=0.45\columnwidth]{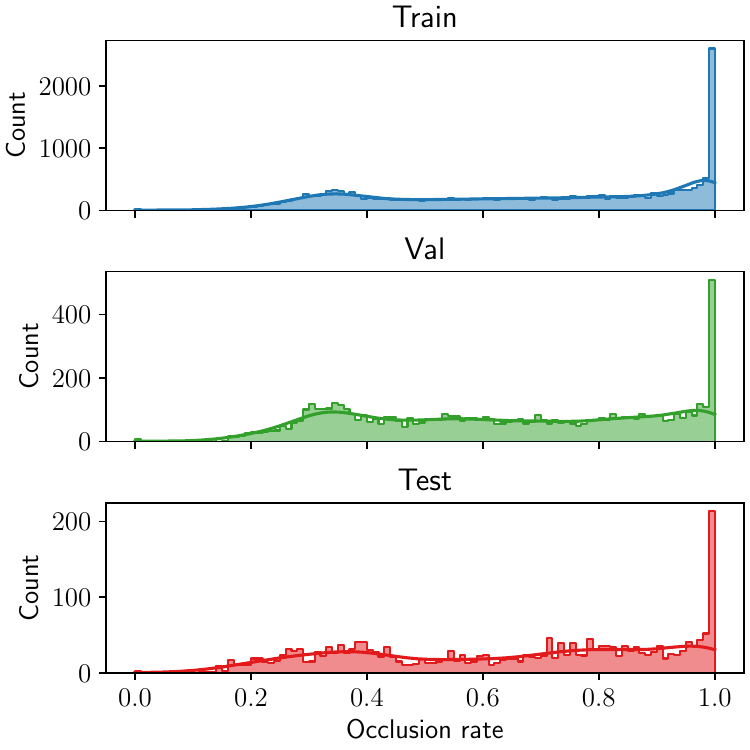}
        \label{fig:original-visibility}
    }
    \caption{Original data distribution.}
    \label{fig:mm:original-distribution}
\end{figure}

These depth and RGB images were combined with the labels transformed to each image and stored as the dataset containing the original images. This resulted in a dataset containing around 800 images per tree, combining to a total of 10,757 images and 28,191 labels of fruits. 
Fig. \ref{fig:mm:original-distribution} shows details on the distribution of fruit orientations and occlusions for the training, validation and testing splits.

\subsubsection{Rendered images dataset}
\begin{figure}[t]
    \centering
    \subfloat[Side view]{
        \includegraphics[width=0.43\columnwidth]{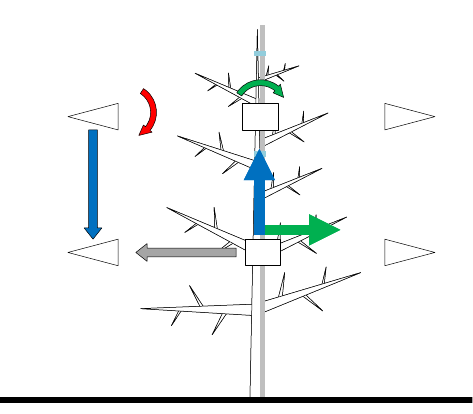}
        \label{fig:orchard-axis-side}
    }
    \hfill
    \subfloat[Top view]{
        \includegraphics[width=0.43\columnwidth]{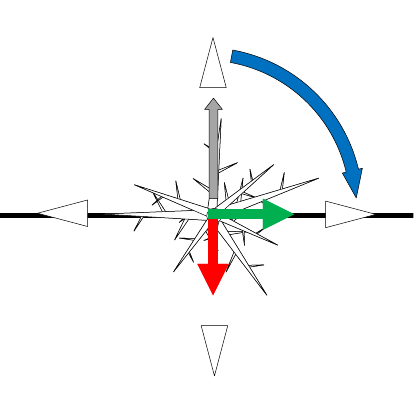}
        \label{fig:orchard-axis-top}
    }
    \caption{Figure showing the camera movement for rendering images around a tree. The wireframe at the center of the tree indicates the origin. 
    Camera roll rotation is shown with a curved red arrow, pitch rotation with a curved green arrow, and yaw rotation with a curved blue arrow.
    Camera height translation is shown with a straight blue arrow, and distance from tree translation is shown with a straight gray arrow.}
    \label{fig:orchard-axis}
\end{figure}

When rendering additional images, any camera extrinsics and intrinsics can be used. In order to keep the images similar to the original images, the same camera intrinsics as used for the patched images were used. As the images can be rendered for each tree, a similar pattern was used for each tree, rotating the camera around the tree at several heights and distances from the tree while adjusting roll and pitch at each camera position. Figure \ref{fig:orchard-axis} shows the adjustment of the camera pose around the tree origin. The tree origin is shown with the wire frame, with the X-axis represented with red, the Y-axis with green and the Z-axis with blue. 
First, the camera was translated to the correct height. 
Next, the camera was rotated around the roll, pitch, and yaw axes to the desired orientation. 
Lastly, it was moved backwards along the rotated X-axis to the desired distance from the tree.
Table \ref{tab:rendering-settings} shows the number of steps and range for each camera pose setting.

\begin{table}[]
    \centering
    \caption{Settings for rendering images around a tree. Yaw rotation and height translation are shown in blue. Camera roll is shown in green and camera pitch is shown in red.}
    \begin{tabular}{rll}
        Setting & Steps & Range \\
        \midrule
        Height & \(3\) &  \(-0.5\) to \(0.7\) m \\
        Roll & \(7\) & \(-\frac{1}{2}\pi\) to \(\frac{1}{2}\pi\) radians \\
        Pitch & \(3\) & \(-\frac{1}{4}\pi\) to \(\frac{1}{4}\pi\) radians \\
        Yaw & \(32\) & \(0\) to \(2\pi\) radians \\
        Distance from tree & \(2\)  & \(2.7\) to \(3.2\) m \\
    \end{tabular}
    \label{tab:rendering-settings}
\end{table}

\begin{figure}[t]
    \centering
    \subfloat[Train orientations]{
        \includegraphics[width=0.3\columnwidth]{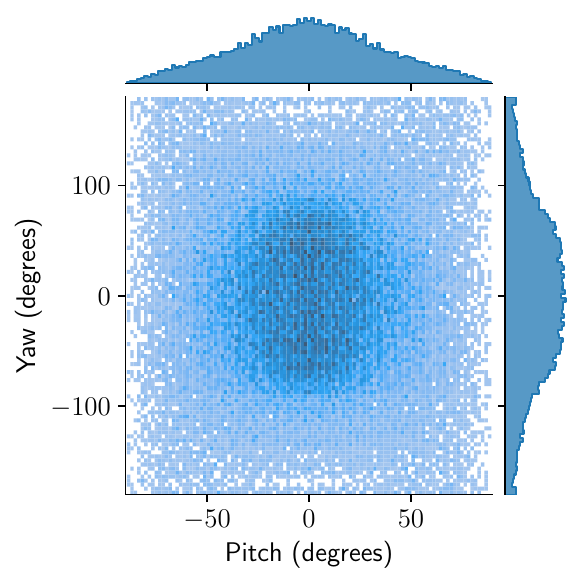}
        \label{fig:rendered-orientation-train}
    }
    \hfill
    \subfloat[Validation orientations]{
        \includegraphics[width=0.3\columnwidth]{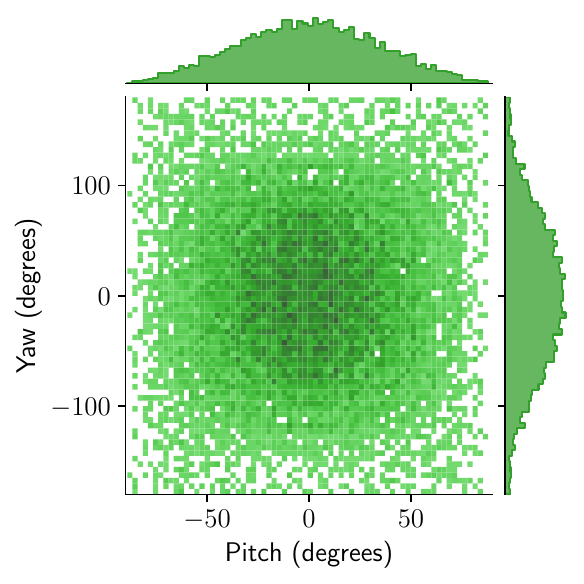}
        \label{fig:rendered-orientation-val}
    }
    \hfill
    \subfloat[Test orientations]{
        \includegraphics[width=0.3\columnwidth]{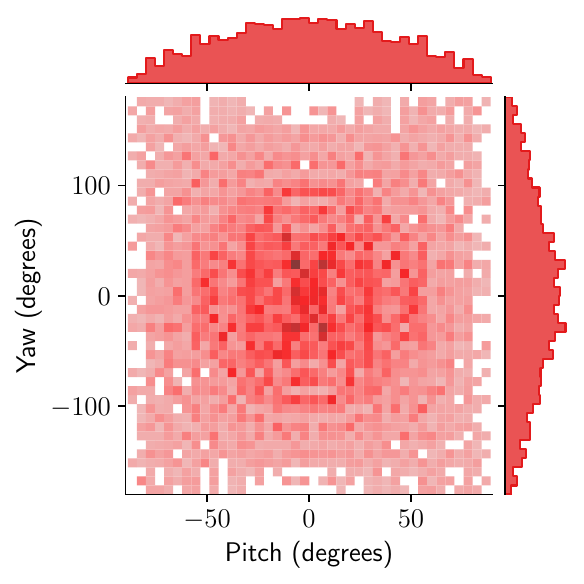}
        \label{fig:rendered-orientation-test}
    }
    \hfill
    \subfloat[3D view of orientations]{
        \includegraphics[width=0.45\columnwidth]{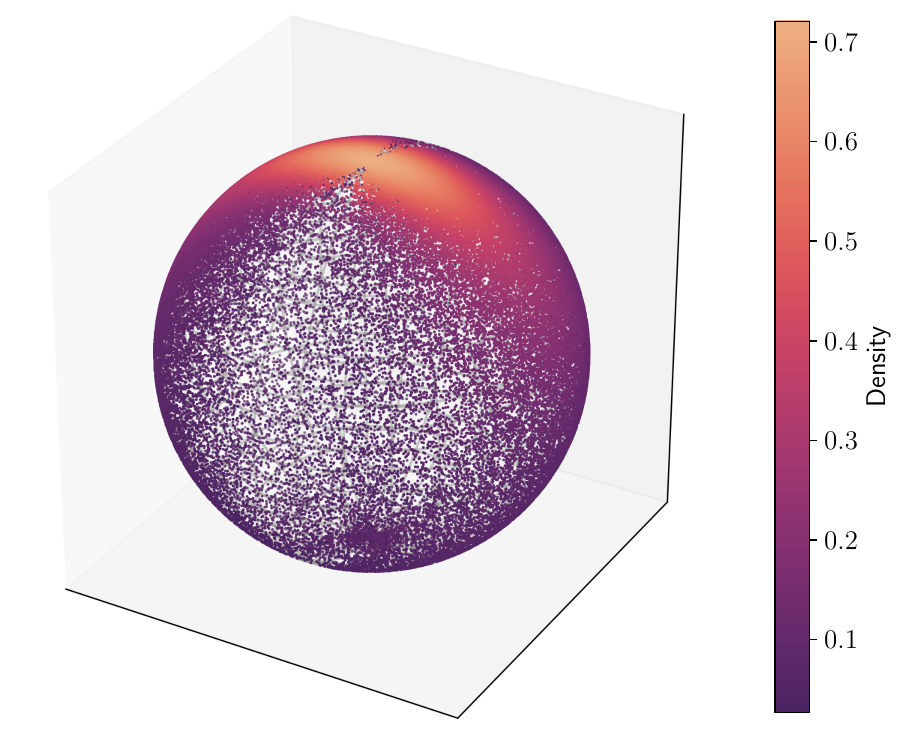}
        \label{fig:rendered-orientation-3d}
    }
    \hfill
    \subfloat[Visibility]{
        \includegraphics[width=0.45\columnwidth]{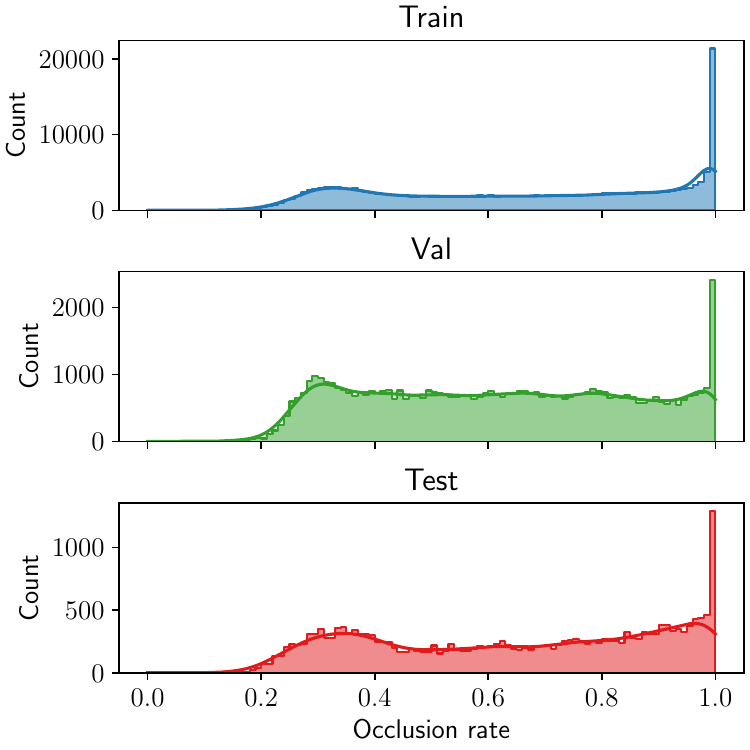}
        \label{fig:rendered-visibility}
    }
    \caption{rendered data distribution.}
    \label{fig:mm:rendered-distribution}
\end{figure}

From each camera pose, an RGB and a depth image were created. These were combined with the labels transformed to each image and stored as the dataset containing rendered images. This resulted in a dataset containing 4,032 images per tree, combining to a total of 52,416 images and 270,732 labels. 
Fig. \ref{fig:mm:rendered-distribution} shows details on the distribution of fruit orientations and occlusions for the training, validation and testing splits.

\subsection{Experiments}\label{sec:mm:exp}
With these datasets, we designed two experiments. First, a label occlusion experiment was set up, described in section \ref{sec:mm:exp:occlusion}. Next, a dataset size experiment was set up, described in section \ref{sec:mm:exp:dataset}. 

\subsubsection{Label occlusion experiment}\label{sec:mm:exp:occlusion}
With this experiment, we aim to determine what occlusion rate in the training labels results in the best model performance. 
Because the fruits are labeled in the 3DGS, even heavily occluded fruits can be provided with labels when training the pose estimation algorithm, and the occlusion rate of each label is known, as described in section \ref{sec:mm:annotating}. In practice, it is desired to detect all fruits, including heavily occluded fruits. However, heavily occluded fruits can be too challenging for a pose estimation algorithm to successfully learn pose estimation. The opposite is also true, as clearly visible fruits without a label can be detected, resulting in false positives. The examples in Fig. \ref{fig:mm:examples-visibility} show different occlusion rates for fruits. At 98\%, the fruit could easily be missed, both by the annotator and the network. However, at 60\%, the detection of the fruit is relatively easy, but orientation estimation remained challenging, as the calyx and peduncle are not visible.
Therefore, we designed an experiment to evaluate the effect of occlusion rate on the model performance.
In addition, model performance is evaluated on a dataset containing labels with a certain occlusion rate. Therefore, we evaluated each model multiple times to determine the effect of occlusion rate not only in model training but also during model evaluation. 

This resulted in training the model with different upper limits of the occlusion rate. This limit was set at 100\%, 95\%, 85\%, 75\%, 65\% and 55\%. For this experiment, the model was either trained and tested on the original images or on rendered images. As the dataset with rendered images is much larger than the dataset with original images, the rendered dataset was subsampled to contain the same number of labels as the original dataset. 
Next, the validation labels were also selected with different upper limits of the occlusion rate. This limit was set at 100\%, 99.9\%, 99\%, 95\%, 90\%, 85\%, 80\%, 75\%, 70\%, 65\%, 60\%, 55\% and 50\%.

The models were evaluated on the F1 score, a commonly used metric for object detection. The F1 score was calculated using the intersection over union (IoU) between the ground truth 3D bounding box and the predicted 3D bounding box, using an IoU threshold of 0.5. If the threshold was above 0.5, the sample was a true positive (TP). Else, it was a false positive (FP). A ground truth box without a predicted box was counted as a false negative (FN). 

\paragraph{Neutral F1 score}
Changing the upper limit of the occlusion rate changed which ground truth labels were present. If the pose estimation method was able to detect a fruit that was not provided with a label, this would be counted as an FP, despite it being a correct prediction. 
This effect influences the precision. Therefore, we calculated the precision using all labels, removing the effect of predicting fruits without labels based on their occlusion.
Then, the F1 score can be calculated using this recalculated precision and the original recall, resulting in a neutralized F1 score. 
This neutralized F1 score gives a more realistic representation of the detection performance, as the metric is not reduced for detecting fruits that are present but were not given a label. 

\subsubsection{Dataset size experiment}\label{sec:mm:exp:dataset}
With this experiment, we aimed to determine the effect of dataset size on the performance. In general, more input data improves the performance of neural networks. However, the number of original images is limited, as annotating these images takes effort. Our proposed pipeline can render additional images, without requiring additional annotation effort. In addition, any camera pose can be used, increasing the variation of fruit poses. 

Our experiment was to change the dataset size and test the effect on apple detection and pose estimation. In this experiment, the models were trained with the best max occlusion rate, as found in the experiment described in \ref{sec:mm:exp:occlusion}. The complete training sets were randomly sampled to create the datasets.
As the rendered images had a different amount of apples present per image than the original images, the dataset sizes were equalized using the number of instances in the dataset relative to the number of instances in the dataset containing the original images. 
Using the original images, the model was trained with datasets containing 12.5\%, 25\%, 50\%, and 100\% of the number of instances. 
Using rendered images, the model was trained with datasets containing 12.5\%, 25\%, 50\%, and 100\% of the number of instances. 
Lastly, the datasets of original images and rendered images were mixed. Using mixed images, the model was trained with datasets containing 25\%, 50\%, 100\%, and 200\% of the number of instances. At these sizes, both sources accounted for half the training instances. At 200\%, all original images were used. The dataset was further expanded using only rendered images, resulting in datasets containing 300\%, 400\%, and 500\% of the number of instances.

Each model was evaluated on the original images test set. The models were evaluated using the F1 score and the neutralized F1 score. In addition, the pose estimation was evaluated. The fruit position was evaluated by looking at the Euclidean distance between the center of the ground truth 3D bounding box and that of the predicted 3D bounding box, as shown in equation \ref{eq:euclid-dist}. The fruit orientation was evaluated by calculating the angular difference between the fruit pose vector of the ground truth instance and that of the predicted instance, as shown in equation \ref{eq:vec-angle}.

\begin{equation}\label{eq:euclid-dist}
    d\left(p, \hat{p}\right) = \sqrt{\left(p_{x} - \hat{p}_x\right)^2 + \left(p_{y} - \hat{p}_y\right)^2 + \left(p_{z} - \hat{p}_z\right)^2}
\end{equation}

\begin{equation}\label{eq:vec-angle}
    \theta\left(v, \hat{v}\right) = \arccos \left( v \cdot \hat{v} \right)
\end{equation}

\section{Results}\label{sec:results}
In this section, the results of the experiments are presented. In section \ref{sec:results:label-occlusion}, the results of the label occlusion experiment are presented. In section \ref{sec:results:dataset-size}, the results of the dataset size experiment are presented.

\subsection{Label occlusion experiment}\label{sec:results:label-occlusion}
\begin{figure}
    \centering
    \subfloat[F1 score]{
        \includegraphics[width=0.43\columnwidth]{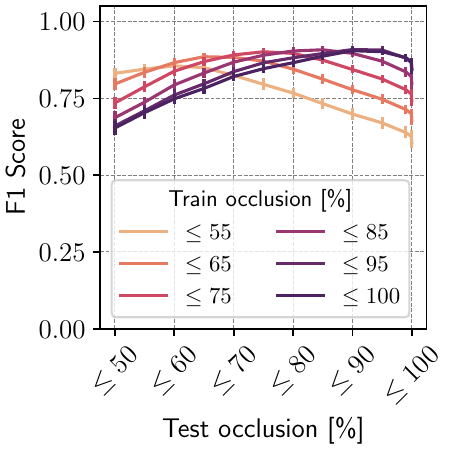}
        \label{fig:real-f1-occlusion}
    }
    \hfill
    \subfloat[Neutral F1 score]{
        \includegraphics[width=0.43\columnwidth]{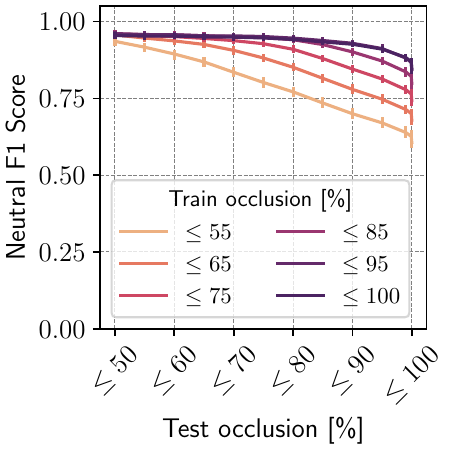}
        \label{fig:real-neutralf1-occlusion}
    }
    \caption{Performance of each train setting on each test setting, using the original images as train and test dataset. Vertical bars indicate the 95\%-interval determined using bootstrapping.}
    \label{fig:real-occlusion}
\end{figure}

\begin{figure}
    \centering
    \subfloat[F1 score]{
        \includegraphics[width=0.43\columnwidth]{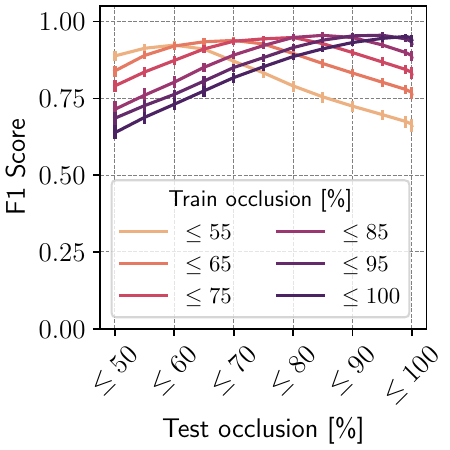}
        \label{fig:gs-f1-occlusion}
    }
    \hfill
    \subfloat[Neutral F1 score]{
        \includegraphics[width=0.43\columnwidth]{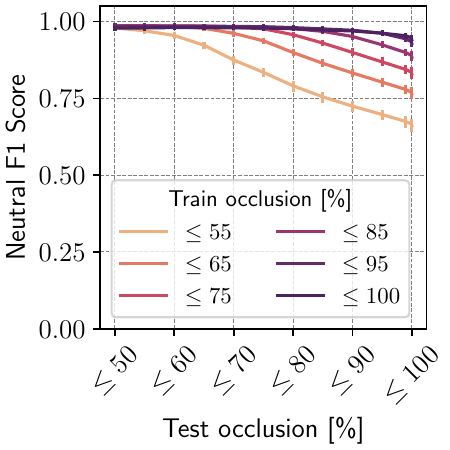}
        \label{fig:gs-neutralf1-occlusion}
    }
    \caption{Performance of each train setting on each test setting, using the rendered images as the train and test dataset. Vertical bars indicate the 95\%-interval determined using bootstrapping.}
    \label{fig:gs-occlusion}
\end{figure}

Through our novel pipeline, we were able to annotate even the heavily occluded fruits, something that is impossible in the conventional way of annotation. This raises the question of what the maximum occlusion rate is for training of the object detector and pose estimator. This experiment, therefore, aimed to determine what occlusion rate in the training labels results in the best model performance. 
Fig. \ref{fig:real-occlusion} and Fig. \ref{fig:gs-occlusion} show the detection performance for the real and rendered images respectively, measured using the F1 and neutralized F1 score, as a function of the upper limit of the occlusion rate in the test dataset, with separate lines for each upper limit of the occlusion rate in the train dataset.

Fig. \ref{fig:real-occlusion} shows the performance when training and testing on the original images. 
With a high test occlusion rate, the F1 and neutral F1 scores are relatively low, as there is a high amount of false negatives, as fruits that are heavily occluded have a label but are not detected.
With a low test occlusion rate, the F1 score (Fig. \ref{fig:real-f1-occlusion}) was relatively low, as there is a high amount of false positives because fruits that are relatively visible have no label but are detected. In contrast, the neutral F1 score, shown in Fig. \ref{fig:real-neutralf1-occlusion}, increased when lowering the test occlusion rate, as the detection of occluded fruits was not counted as false positives, better reflecting the ability to detect the fruits present. 
For each of the different train occlusion rates, the highest F1 score was achieved when it resembled the test occlusion rate. This indicates that the neural network not only learned to detect the fruits but also learned at what maximum level of occlusion they should be detected.
Training with an occlusion rate of \(\leq95\%\) or \(\leq100\%\), using all fruit instances, resulted in the highest neutral F1 score, but the differences between these two train settings are small, which further reduced when lowering the test occlusion rate. This shows that providing labels with a higher occlusion rate adds value up to \(95\%\). Adding the last \(5\%\) of the most occluded fruits did not have a significant effect on the performance of the model. 

Fig. \ref{fig:gs-occlusion} shows the performance when training and testing on the dataset containing rendered images. Here, the same effects can be observed as when using the original images. However, the F1 score and neutral F1 score are slightly higher compared to when using the original images, indicating that the detection of fruits in the rendered images is somewhat easier than in the original images. Using rendered images, there was a smaller reduction in performance in the case of high test occlusion rates, compared to using the original images. 

In order to better analyze the performance, the best train occlusion rate for each test occlusion rate was determined. Table \ref{tab:occlusion-rank-test} shows the train occlusion rate that achieved the best neutral F1 score on each test occlusion rate, for the rendered and original images. If there were no significant differences between multiple train occlusion rates for a given test occlusion rate, all train occlusion rates in the best performing group are shown. The significant groups were calculated using Tukey's HSD test. 
Here, it can be seen that as the test occlusion rate is reduced, the best train occlusion rate was lower as well, although not as much. This effect was greater in the rendered images than in the original images. 
From the table, it can be seen that a train occlusion rate of \(\leq100\%\) performs best when the test occlusion rate is also high. From a test occlusion rate of \(\leq95\%\), a train occlusion rate of \(\leq95\%\) starts to perform best. This shows that annotating the \(5\%\) of most occluded fruits only matters when it is important to detect these fruits. However, when only the least occluded fruits need to be detected, it is beneficial to annotate more occluded fruits, which can be observed at low test occlusion rates, where the best train occlusion rate is \(\leq75\%\) or \(\leq85\%\), for the rendered and original images respectively. The difference between the best train occlusion rate for a given test occlusion rate is bigger when using the original images than when using the rendered images. When using the original images, a train occlusion rate of \(\leq95\%\) was in the best performing group still when the test occlusion rate was \(\leq55\%\). 
To determine which train setting performed the best across all test settings, the number of occurrences in the group that performed best was counted for all levels of the test occlusion rate. This resulted in the highest count for models trained with an occlusion rate of \(\leq95\%\), which occurred in 7 of the 13 test settings when using the rendered images and in 9 of 13 test settings when using the original images. 

\begin{table}
    \centering
    \caption{Table showing the group of train settings that performed best on each test setting. Multiple train settings are shown if \(p>0.05\). }
    \begin{tabular}{r|ll}
        \multirow{2}{*}{Test occlusion [\%]} & \multicolumn{2}{c}{Train occlusion [\%]} \\
        & Rendered & Original \\
        \midrule
        \(\leq100\) & \(\leq100\) & \(\leq100\) \\
        \(\leq99.9\) & \(\leq100\) & \(\leq100\) \\
        \(\leq99\) & \(\leq100\) & \(\leq100\) \\
        \(\leq95\) & \(\leq100\),   \(\leq95\) & \(\leq95\) \\
        \(\leq90\) & \(\leq95\) & \(\leq95\) \\
        \(\leq85\) & \(\leq95\) & \(\leq95\) \\
        \(\leq80\) & \(\leq95\) & \(\leq95\) \\
        \(\leq75\) & \(\leq95\) & \(\leq95\) \\
        \(\leq70\) & \(\leq95\) & \(\leq95\) \\
        \(\leq65\) & \(\leq95\), \(\leq75\) & \(\leq85\), \(\leq95\) \\
        \(\leq60\) & \(\leq75\) & \(\leq95\), \(\leq85\) \\
        \(\leq55\) & \(\leq75\), \(\leq65\) & \(\leq85\), \(\leq95\) \\
        \(\leq50\) & \(\leq75\) & \(\leq85\) \\
    \end{tabular}
    \label{tab:occlusion-rank-test}
\end{table}

In Table \ref{tab:train-performance}, the performance when averaging across all test settings is shown for each train setting. The neutral F1 score is calculated by averaging the performance across all test settings, and significant differences are determined using Tukey's HSD test. The rank is determined by ordering the train settings on the neutral F1 score, using Tukey's HSD test to assign the same rank if there are no significant differences between train settings. 
In this table, a similar trend in the results was visible when using the original or rendered images. It can be seen that the models trained with an occlusion rate of \(\leq95\%\) and \(\leq100\%\) achieved similar neutral F1 scores and outperformed all other train settings. When looking at the average rank, training with an occlusion rate of \(\leq95\%\) was the best, achieving nearly 1 rank higher on average than the next best train setting. The models trained with \(\leq100\%\) and \(\leq85\%\) are very similar in average rank, but there is a significant difference in neutral F1 score. This difference can be attributed to the large differences in neutral F1 score at high test max occlusion rates. Here, the models trained with an occlusion rate of \(\leq100\%\) achieved a relatively large difference in neutralized F1 score, while the ranking did not take the absolute difference into account. 

\begin{table}
    \centering
    \caption{Table showing the neutralized F1 score and average rank of train settings across all test settings. Significantly different groups are indicated with different letters if \(p<0.05\). The best performance is highlighted in bold.}
    \begin{tabular}{r|llll}
        \multirow{2}{*}{Train occlusion [\%]} & \multicolumn{2}{c}{Rendered} & \multicolumn{2}{c}{Original} \\
        & Neutral F1 & Rank & Neutral F1 & Rank \\
        \midrule
        \(\leq100\) & \textbf{0.969\textsuperscript{a}} & 2.53 & \textbf{0.924\textsuperscript{a}} & 2.15 \\
        \(\leq95\) & \textbf{0.970\textsuperscript{a}} & \textbf{1.54} & \textbf{0.927\textsuperscript{a}} & \textbf{1.31} \\
        \(\leq85\) & 0.954\textsuperscript{b} & 2.46 & 0.910\textsuperscript{b} & 2.23 \\
        \(\leq75\) & 0.928\textsuperscript{c} & 2.92 & 0.878\textsuperscript{c} & 3.61 \\
        \(\leq65\) & 0.888\textsuperscript{d} & 4.15 & 0.834\textsuperscript{d} & 4.61 \\
        \(\leq55\) & 0.808\textsuperscript{e} & 5.46 & 0.769\textsuperscript{e} & 5.69 \\
    \end{tabular}
    \label{tab:train-performance}
\end{table}

\begin{figure}
    \centering
    \subfloat[Recall]{
        \includegraphics[width=0.3\columnwidth]{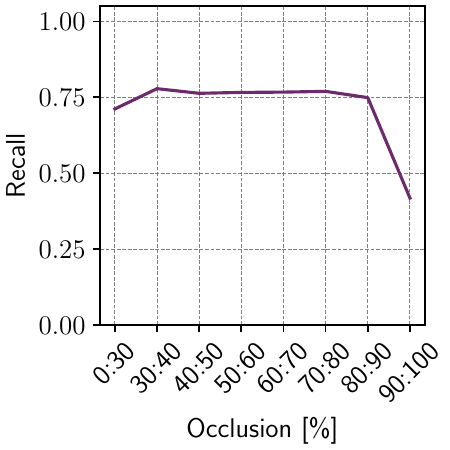}
        \label{fig:original-recall-occlusion}
    }
    \hfill
    \subfloat[Position error]{
        \includegraphics[width=0.3\columnwidth]{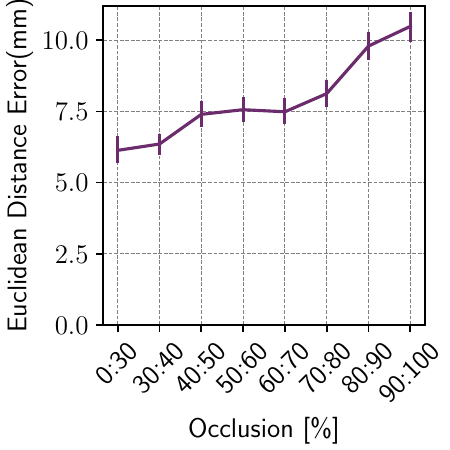}
        \label{fig:original-position-occlusion}
    }
    \hfill
    \subfloat[Orientation error]{
        \includegraphics[width=0.3\columnwidth]{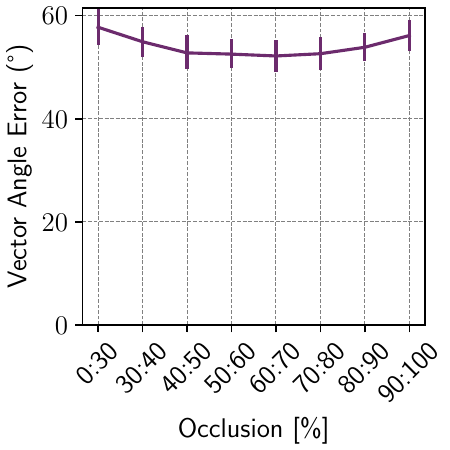}
        \label{fig:original-orientation-occlusion}
    }
    \caption{Performance as a function of different levels of fruit occlusion, using the original images as the train and test dataset. Vertical bars indicate the 95\% confidence interval.}
    \label{fig:original-occlusion-bins}
\end{figure}

\begin{figure}
    \centering
    \subfloat[Recall]{
        \includegraphics[width=0.3\columnwidth]{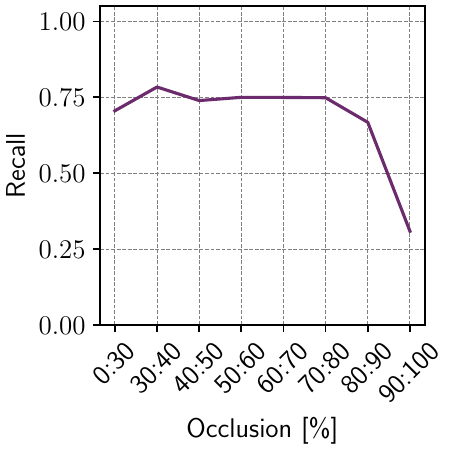}
        \label{fig:rendered-recall-occlusion}
    }
    \hfill
    \subfloat[Position error]{
        \includegraphics[width=0.3\columnwidth]{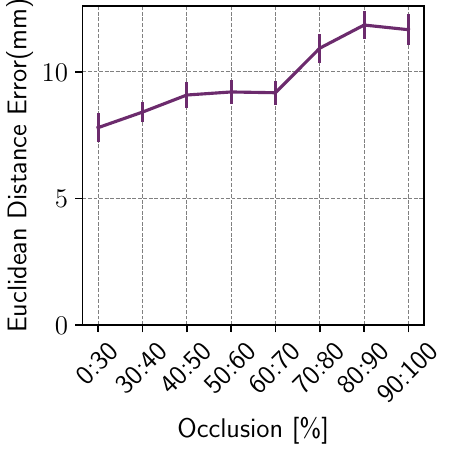}
        \label{fig:rendered-position-occlusion}
    }
    \hfill
    \subfloat[Orientation error]{
        \includegraphics[width=0.3\columnwidth]{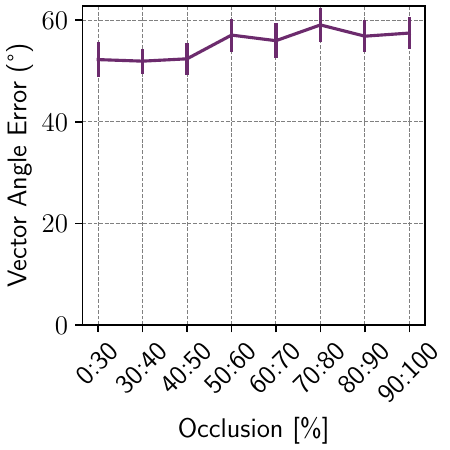}
        \label{fig:rendered-orientation-occlusion}
    }
    \caption{Performance as a function of different levels of fruit occlusion, using the rendered images as the train dataset and the original images as the test dataset. Vertical bars indicate the 95\% confidence interval.}
    \label{fig:rendered-occlusion-bins}
\end{figure}

As training with an occlusion rate of \(\leq95\%\) resulted in the best performance, we used these models to analyze the performance for multiple levels of occlusion, by dividing the test instances into bins. Eight bins were created, in steps of 10\% from fully occluded. Fruits with an occlusion rate of less than 30\% were contained in a single bin, as there were relatively few fruits with these low occlusion rates. 
Fig. \ref{fig:original-occlusion-bins} shows the performance as a function of the occlusion rate of instances in the test dataset. The models were trained on a dataset of original images with an occlusion rate of \(\leq95\%\) and tested on the original images. 
The detection performance is shown with the recall, shown in Fig. \ref{fig:original-recall-occlusion}. In this case, the performance remained consistent as the occlusion rate increased. There was a large decrease in performance for the bin with the most occluded fruits, indicating that a large amount of these fruits were not detected. 
The position estimation performance is shown with the Euclidean distance, shown in Fig. \ref{fig:original-position-occlusion}. In this case, it can be seen that the position estimation accuracy reduced when the fruits became more occluded. 
The orientation estimation performance is shown with the vector angle error, shown in Fig. \ref{fig:original-orientation-occlusion}. There was minimal effect of the occlusion rate on the orientation estimation. For the least and most occluded fruits, the orientation estimation was slightly worse.
Fig. \ref{fig:rendered-occlusion-bins} shows the performance as a function of the occlusion rate of instances in the test dataset. The models were trained on a dataset of rendered images with an occlusion rate of \(\leq95\%\) and tested on the original images.
The detection performance is shown with the recall, shown in Fig. \ref{fig:rendered-recall-occlusion}. In this case, the performance remained consistent as the occlusion rate increased. There was a large decrease in performance for the bin with the most occluded fruits, indicating that a large amount of these fruits were not detected. 
The position estimation performance is shown with the Euclidean distance, shown in Fig. \ref{fig:rendered-position-occlusion}. In this case, it can be seen that the position estimation accuracy reduced when the fruits became more occluded. 
The orientation estimation performance is shown with the vector angle error, shown in Fig. \ref{fig:rendered-orientation-occlusion}. There was minimal effect of the occlusion rate on the orientation estimation. The orientation estimation became slightly worse when the fruits became more occluded.

\subsection{Dataset size experiment}\label{sec:results:dataset-size}
\begin{figure}[th]
    \centering
    \subfloat[F1 score]{
        \includegraphics[width=0.43\columnwidth]{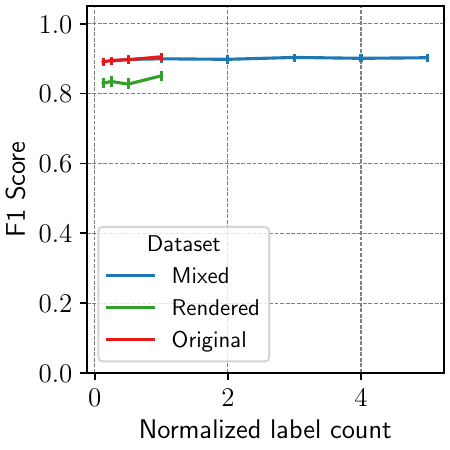}
        \label{fig:dataset-f1-labelcount}
    }
    \hfill
    \subfloat[Neutral F1 score]{
        \includegraphics[width=0.43\columnwidth]{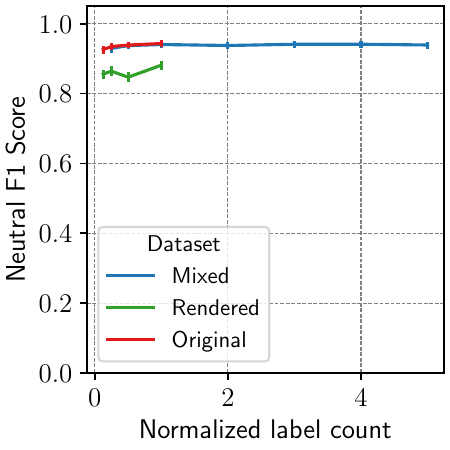}
        \label{fig:dataset-neutralf1-labelcount}
    }
    \caption{Detection performance as a function of the number of labeled apple instances. Different colors indicate the type of images used in the train dataset. The number of labeled apple instances was normalized relative to the number of labeled apple instances in the original images, which was 15,419.
    Vertical bars indicate the 95\%-interval determined using bootstrapping.}
    \label{fig:dataset-detection-labelcount}
\end{figure}

With this experiment, we aimed to determine the effect of dataset size on the performance. For this experiment, multiple training datasets were evaluated on the test split of the dataset consisting of the original images. 
Fig. \ref{fig:dataset-detection-labelcount} shows the detection performance as a function of the number of labeled apple instances in the training set for original, rendered and mixed training sets. The models were trained on a dataset with an occlusion rate of \(\leq95\%\) and tested on a dataset containing only the original images, with an occlusion rate of \(\leq85\%\). Here it can be seen that the models trained on just rendered data perform a lot worse than the models trained on the original data. Using mixed data resulted in similar performance to using the original images. Expanding the dataset beyond the amount of instances in the dataset containing original images by adding more rendered images had no significant effect on the detection performance. 

\begin{figure}[th]
    \centering
    \subfloat[Position error]{
        \includegraphics[width=0.43\columnwidth]{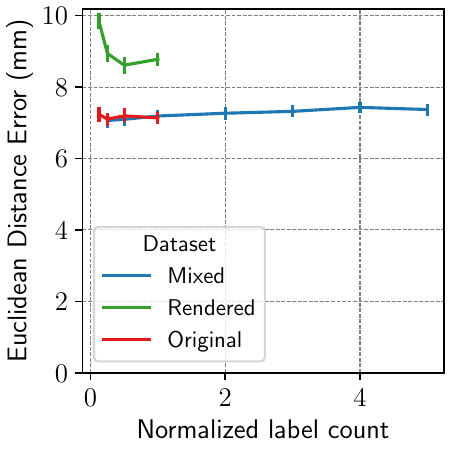}
        \label{fig:position-labelcount}
    }
    \hfill
    \subfloat[Orientation error]{
        \includegraphics[width=0.43\columnwidth]{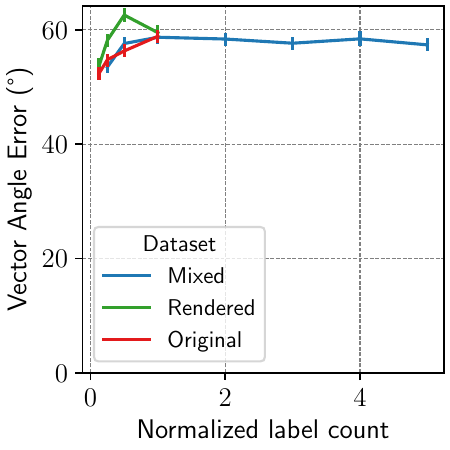}
        \label{fig:orientation-labelcount}
    }
    \caption{Pose estimation performance as a function of the number of labeled apple instances. Different colors indicate the type of images used in the train dataset. The number of labeled apple instances was normalized relative to the number of labeled apple instances in the original images, which was 15,419.
    Vertical bars indicate the 95\%-interval determined using bootstrapping.}
    \label{fig:pose-labelcount}
\end{figure}

\begin{figure}
    \centering
    \subfloat[Normalized label count of 0.125]{
        \includegraphics[width=0.43\columnwidth]{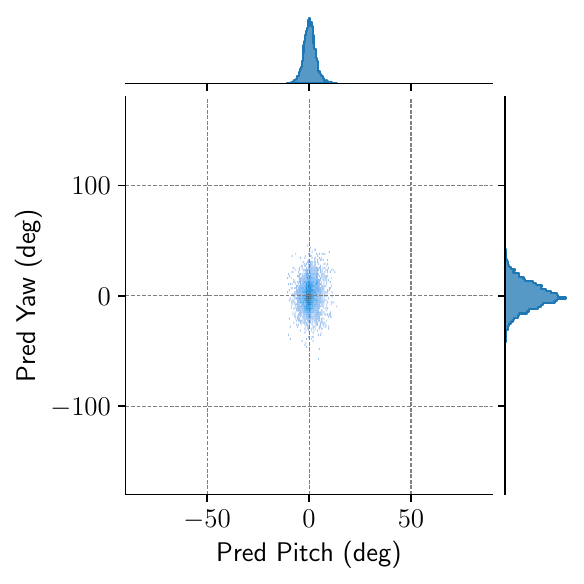}
        \label{fig:predicted-orientations-0125}
    }
    \hfill
    \subfloat[Normalized label count of 0.25]{
        \includegraphics[width=0.43\columnwidth]{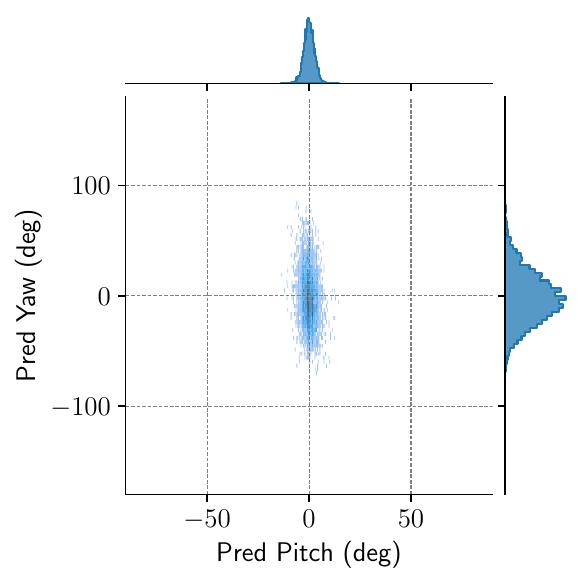}
        \label{fig:predicted-orientations-0250}
    }
    \hfill
    \subfloat[Normalized label count of 0.5]{
        \includegraphics[width=0.43\columnwidth]{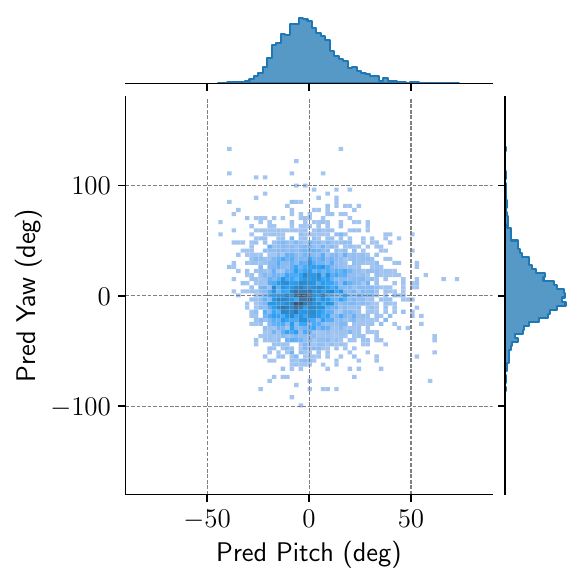}
        \label{fig:predicted-orientations-0500}
    }
    \hfill
    \subfloat[Normalized label count of 1.0]{
        \includegraphics[width=0.43\columnwidth]{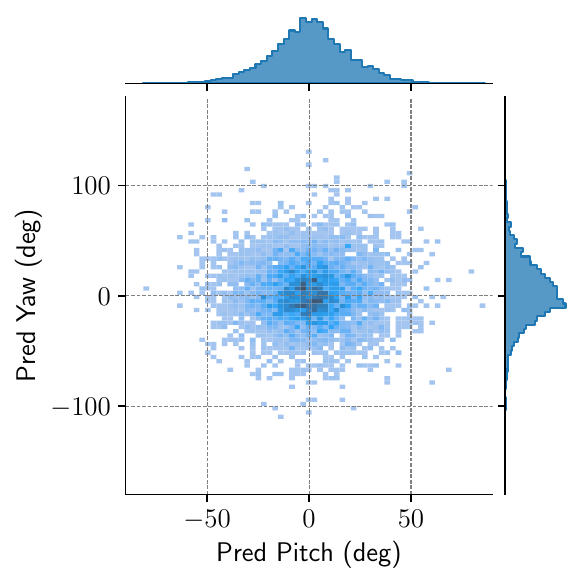}
        \label{fig:predicted-orientations-1000}
    }
    \caption{Figure showing the orientations predicted for different training dataset sizes, using only the original images. The test dataset consisted of the original images. The distribution of orientations in this split is shown in Fig. \ref{fig:original-orientation-test}.}
    \label{fig:original-predicted-orientations}
\end{figure}

Fig. \ref{fig:pose-labelcount} shows the pose estimation performance as a function of the number of labeled apple instances in the training set for original, rendered and mixed training sets. The models were trained on a dataset with an occlusion rate of \(\leq95\%\) and tested on a dataset containing only the original images, with an occlusion rate of \(\leq85\%\). 
For the position estimation, shown in Fig. \ref{fig:position-labelcount}, there was no significant effect of the dataset size on position estimation accuracy when using the dataset containing the original images or mixed images. When using the dataset containing the rendered images, there was a significant improvement in position estimation accuracy when increasing the dataset size. 
For the orientation estimation, shown in Fig. \ref{fig:orientation-labelcount}, the smallest datasets resulted in the best orientation estimation accuracy, while the opposite was expected. Increasing the dataset size up to a normalized label count of 1.0 reduced the orientation estimation accuracy, which stabilized for normalized label counts greater than 1.0. 
To obtain an improved understanding of the reduction in orientation estimation accuracy when the dataset size was increased, the distribution of the estimated orientations for different dataset settings is shown in Fig. \ref{fig:original-predicted-orientations}. Here, it can be observed that the models trained with the smallest datasets do not predict variation in the orientation at all. This corresponds to always predicting no orientation, which can be used to determine whether a model is predicting a correct orientation. While the models trained with larger datasets resulted in more variation in the predicted orientation, this did not improve the orientation estimation. This indicates that the model did not correctly learn to predict the fruit orientation but was likely randomly guessing a fruit orientation. This effect also occurs when training with the datasets containing the original and rendered images, for which the results are shown in \ref{sec:app:orientation-pred}. 

\subsection{Comparison of annotation effort and performance}
Table \ref{tab:model-performance-comparison} shows the annotation effort and performance of multiple apple pose estimation methods. The methods are compared on the number of annotations required to obtain the datasets used for training, the resulting number of instances available for training, the detection performance, and the pose estimation performance. 
Through our proposed pipeline, a fraction of the annotation effort was required, while more instances were in the dataset. While both other methods needed as many annotations as instances in the dataset, our proposed pipeline reduced the annotation effort by 99.6\%, when considering only the original images. This if further reduced when rendering additional images. 
In terms of detection performance, our proposed pipeline improved the detection performance over both compared methods. 
In terms of pose estimation performance, our proposed pipeline performed worse than the multi-stage method presented by \citet{RN205}, which achieved a much lower median angle error. Compared to FRESHNet \cite{RN226} trained on their dataset, our pitch and yaw errors were similar. Both other methods did not report the position accuracy. 

\begin{table}[H]
    \centering
    \caption{Table showing the annotation effort and performance of apple pose estimation methods. Values indicated with \textsuperscript{1} were not directly reported in their related work and were calculated based on their reported precision and recall.}
    \begin{tabular}{l|ccc}
        Model & \citet{RN205} & FRESHNet \cite{RN226} & Ours \\
        \midrule
        \multicolumn{4}{l}{\textit{Annotation effort}} \\
        3D Annotations & N/A & 9,044 & 105 \\
        2D Annotations & 1,000 & N/A & N/A \\
        Instances in dataset & 1,000 & 9,044 & 28,191 \\
        \midrule
        \multicolumn{4}{l}{\textit{Detection performance}} \\
        mAP & 0.8911 & 0.782 & -- \\
        F1 & 0.901\textsuperscript{1} & 0.784\textsuperscript{1} & 0.943 \\
        \midrule
        \multicolumn{4}{l}{\textit{Pose estimation performance}} \\
        Pitch Error & -- & 23.88° & 25.76° \\
        Yaw Error & -- & 34.43° & 32.99° \\
        Vector Angle Error & 17.6° & -- & 48.1° \\
        Euclidean Distance & -- & -- & 7.13 mm \\
    \end{tabular}
    \label{tab:model-performance-comparison}
\end{table}

\section{Discussion}\label{sec:disc}

\subsection{Discussion of the 3D annotation pipeline}\label{sec:disc:3d-annot}
3D scene reconstruction has been used for agricultural tasks, such as fruit detection and segmentation, in related work \cite{RN268,RN265,RN266}. 
These works exploited the reconstruction mainly to generate depth data or to augment datasets.
Our proposed pipeline extends this by using 3DGS to reconstruct an orchard scene, enabling the manual annotation in the 3DGS, projecting the 3D annotations to the 3D image, and allowing the collection of a consistent and large-scale dataset for 5D fruit pose estimation to train and evaluate a pose-estimation method. 

Using the 3DGS, the number of manual annotations was reduced by 99.6\% compared to annotating the data on a 2D image level. Moreover, as the 3DGS gives a more complete view of a fruit compared to individual images, the pose could be annotated more accurately and the projection resulted in consistent annotation of the image data even in situations with low visibility due to occlusions. The pipeline furthermore enabled the calculation of the occlusion rate of the fruit instances in the images. This allowed us to study the effect of occlusions. Through this experiment, we were able to determine that the best performance could be achieved when fruits with an occlusion rate of \(\leq95\%\) were included in the train dataset. 

The occlusion rate of an apple in an image was determined using the 3D point cloud of that apple. However, the point cloud does not completely cover the area of the fruit, leaving small holes, as can be seen in Fig. \ref{fig:mm:global-labels:fruit}. In addition, the depth determined from the 3DGS contains errors if small obstacles were in front of the fruit. These two effects resulted in the estimated occlusion rate of fully visible fruits being approximately 13\% instead of the expected 0\%. However, this affects the occlusion rate of all fruits in a similar way, and does not affect the comparison between different train settings. Therefore, we believe the conclusion remains valid. 

The 3DGS that was created contained a single set of trees. This resulted in a relatively homogeneous dataset, which does not represent the variety in fruit growing systems. This limits the conclusions to similar conditions as in this orchard. Including more variation in the dataset is valuable to improve the generalizability of the pose-estimation method but the value of the pipeline remains, as it simplifies creating new datasets. 

\subsection{Discussion of pose estimation}\label{sec:disc:results}
With our experiments, the effect of adjusting the upper limit of the occlusion rate of the instances in the training set on model performance was shown. For the detection performance, it was found that the highest F1 score can be achieved when the test occlusion rate was similar to the train occlusion rate. 
The trained models learned that apples occluded more than the upper limit should not be detected. 
As detecting a fruit that is actually there but occluded more than the upper limit should not be considered an error by the detection model, we introduced the neutral F1 score. 
This led to the finding that using more occluded fruits in the training data improved the detection performance.

Concerning the performance of pose estimation, fruit occlusions reduced the accuracy of estimating the position, with higher error for more occluded instances. For orientation estimation this relationship was not found. It should be noted, however, that orientation estimation showed a high error between 50° and 60° for all occlusion rates. If instead of using FRESHnet, the orientation angles were always set to 0.0, we obtained a lower orientation error. From this, we have to conclude that FRESHnet was unable to estimate the orientation of the apples. 
Our results are similar to the orientation estimation accuracy reported in \cite{RN226}. While they drew positive conclusions from the results, we believe that we have to conclude that FRESHnet is unable to estimate the orientation of the apples. 

\subsection{Future work}
In this work, we created a 3DGS of a single set of trees. This resulted in a relatively homogeneous dataset, which does not represent the variety in fruit growing systems. Therefore, future work should focus on collecting datasets in different fruit growing systems, allowing for better generalization. 

Although we focused on the evaluation of one pose-estimation method, the proposed pipeline can be used to benchmark more pose estimation methods. Some pose estimation methods define the pose not as an oriented 3D bounding box, but by a set of keypoints, e.g. \citet{RN205}, for which the labels cannot be generated by our pipeline. Therefore, expanding our pipeline to generate other label types is valuable to enable evaluating multiple methods under the same circumstances. 

Given the poor performance of orientation estimation, focus should be on improving the orientation estimation, for instance by tuning the weights, improving the loss function on the orientation, or adjusting the network architecture. 
In addition, the occlusion rate can be incorporated during training, to prevent the model from being punished for detecting fruits with high occlusion rates, that would normally not have gotten a label. 

Given the effect that the occlusion rate had on the performance of the model, a comparison with manually providing labels for each image is interesting. It is not known until what occlusion rate annotators provide labels for fruits, and how the occlusion rate affects the accuracy of the annotation. 
Therefore, it is recommended to research up to what occlusion rate annotators provide annotations and how accurate these annotations are. In addition, it is interesting to research how consistent multiple annotators are, both for our method and for annotating each instance.

\section{Conclusion}\label{sec:conc}
In this study, we presented a pipeline that utilized 3DGS to reconstruct an orchard scene, simplified the manual annotation process, automated projection of the annotations to images, and performed the training and evaluation of a pose-estimation method. To obtain labels for the original images, only 105 annotations were required to obtain 28,191 training labels for 10,757 images, a reduction of 99.6\%. 

Through the 3DGS, the occlusion rate of each fruit in each image was determined. To determine the effect of the occlusion rate on model performance, we trained the model with multiple upper limits for the occlusion rate of labels. It was found that each model performed best when tested on labels with a similar upper limit of occlusion rate. It was found that training with an occlusion rate up to \(\leq95\%\) resulted in the best performance, with a neutral F1 score of 0.927 on the original images and 0.970 on the rendered images. 
Adjusting the size of the training dataset had small effects on the model performance in terms of F1 score and pose-estimation accuracy. It was found that the position of fruits could be determined with an average error of 7.13 mm. The least occluded fruits had the best position estimation, which worsened as the fruits became more occluded. 
It was also found that the tested pose estimation method was unable to correctly learn the orientation estimation of apples, with an average vector angle error of 48.1°, coming from a pitch error of 25.76° and a yaw error of 32.99°.

\section{Acknowledgments}
This work was partially funded by the Dutch Ministry of LVVN (Agriculture, Fisheries, Food Security, and Nature), under the project code KB-38-001-005, and by the Netherlands Organization for Scientific Research (NWO grant 17626, project Synergia).

\newpage

\bibliographystyle{elsarticle-num-names} 
\bibliography{refs}

%% The Appendices part is started with the command \appendix;
%% appendix sections are then done as normal sections
\newpage
\appendix

\section{Camera settings when taking images}\label{sec:app:cam-settings}

\begin{table}[H]
    \centering
    \caption{Settings of camera for taking images}
    \begin{tabular}{rl}
        Setting & Value \\
        \midrule
        Camera type &  Nikon Z6\\
        Lens type & NIKKOR Z 24-70mm f/2.8s\\
        Image size & 6048 by 4024\\
        White balancing & Cloudy, approx. 6000K\\
        F-stop & f/6.3\\
        ISO & 1600\\
        Exposure & 1/160\\
        Focal length & 24.0 mm\\
    \end{tabular}
    \label{tab:cam-settings}
\end{table}

\section{Settings for Structure from Motion}\label{sec:app:sfm-settings}

\begin{table}[H]
    \centering
    \caption{Settings used for camera pose estimation and sparse point cloud generation}
    \begin{tabular}{lll}
        Step & Setting & Value \\
        \midrule
        Camera pose estimation & Accuracy &  High \\
        & Key point limit & 40,000 \\
        & Tie point limit & 4,000 \\
        Point cloud generation & Quality & High \\
        & Depth-filtering & Mild \\
    \end{tabular}
    \label{tab:sfm-settings}
\end{table} 

\section{Settings for training 3DGS}\label{sec:app:3dgs-settings}

\begin{table}[H]
    \centering
    \caption{Settings used for training the 3DGS in the software Jawset Postshot}
    \begin{tabular}{rl}
        Setting & Value \\
        \midrule
        Model profile &  3DGS MCMC \cite{RN272} \\
        Sample images & 3200 pixels \\
        Max number of Gaussians & 7,000,000 \\
        Max training steps & 37,000 \\
        Antialiasing & Disabled \\
        Create Sky Model & Disabled \\
    \end{tabular}
    \label{tab:3dgs-settings}
\end{table}

\section{Additional results on orientation prediction}
\label{sec:app:orientation-pred}

\begin{figure}[H]
    \centering
    \subfloat[Normalized label count of 0.125]{
        \includegraphics[width=0.43\columnwidth]{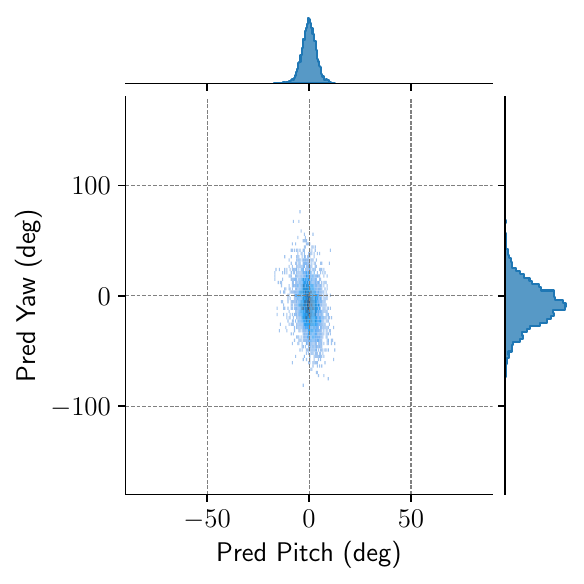}
        \label{fig:app-rendered-predicted-orientations-0125}
    }
    \hfill
    \subfloat[Normalized label count of 0.25]{
        \includegraphics[width=0.43\columnwidth]{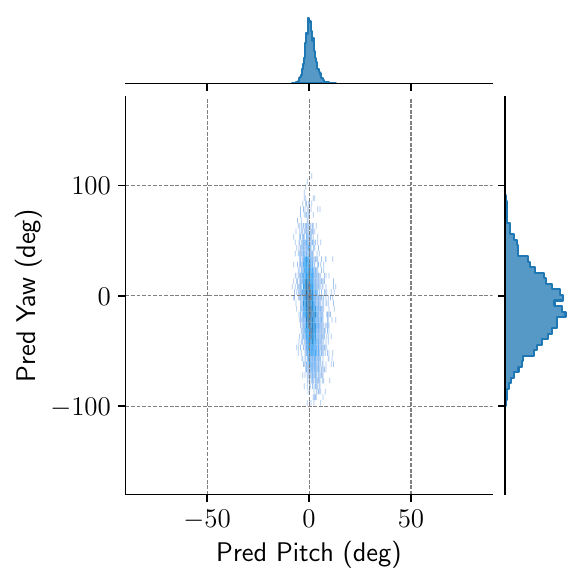}
        \label{fig:app-rendered-predicted-orientations-0250}
    }
    \hfill
    \subfloat[Normalized label count of 0.5]{
        \includegraphics[width=0.43\columnwidth]{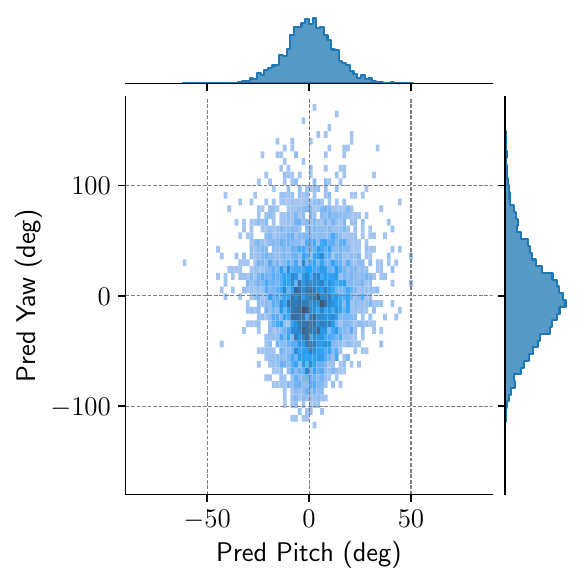}
        \label{fig:app-rendered-predicted-orientations-0500}
    }
    \hfill
    \subfloat[Normalized label count of 1.0]{
        \includegraphics[width=0.43\columnwidth]{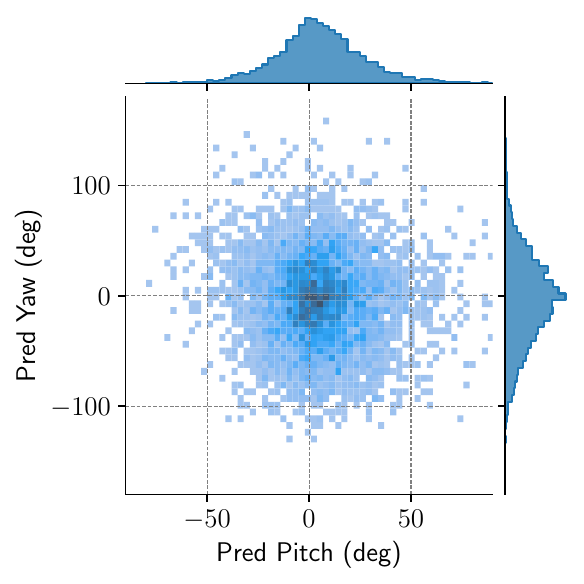}
        \label{fig:app-rendered-predicted-orientations-1000}
    }
    \caption{Figure showing the orientations predicted for different training dataset sizes, using only the rendered images. The test dataset consisted of the original images. The distribution of orientations in this split is shown in Fig. \ref{fig:original-orientation-test}.}
    \label{fig:app-rendered-predicted-orientations}
\end{figure}

\begin{figure}[H]
    \centering
    \subfloat[Normalized label count of 0.25]{
        \includegraphics[width=0.43\columnwidth]{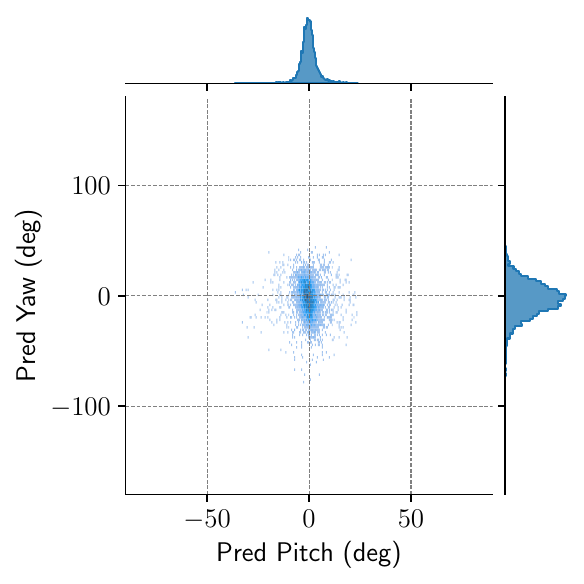}
        \label{fig:app-mixed-predicted-orientations-0250}
    }
    \hfill
    \subfloat[Normalized label count of 0.5]{
        \includegraphics[width=0.43\columnwidth]{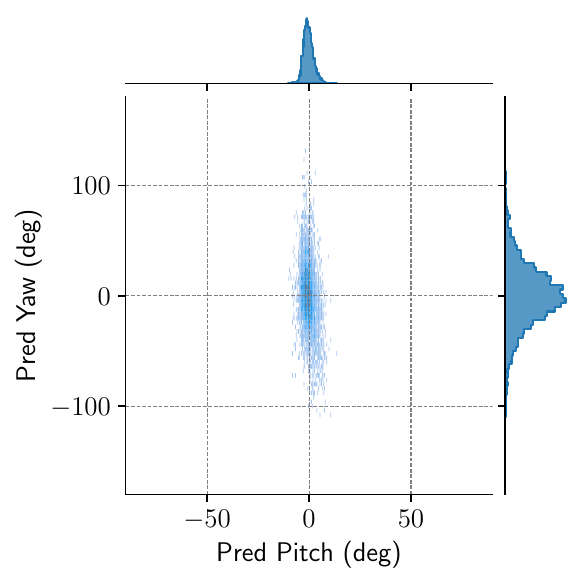}
        \label{fig:app-mixed-predicted-orientations-0500}
    }
    \hfill
    \subfloat[Normalized label count of 1.0]{
        \includegraphics[width=0.43\columnwidth]{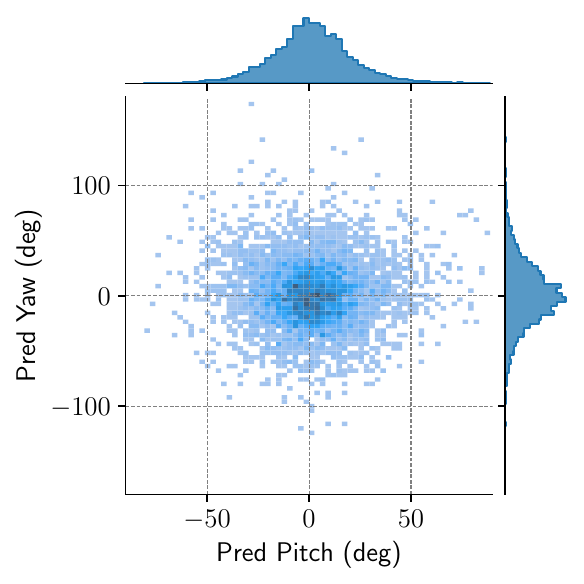}
        \label{fig:app-mixed-predicted-orientations-1000}
    }
    \hfill
    \subfloat[Normalized label count of 2.0]{
        \includegraphics[width=0.43\columnwidth]{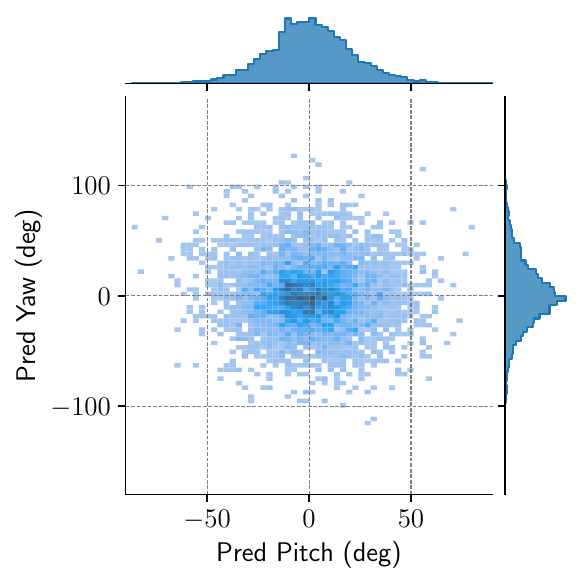}
        \label{fig:app-mixed-predicted-orientations-2000}
    }
    \caption{Figure showing the orientations predicted for different training dataset sizes, using both original and rendered images up to a normalized label count of 2.0. The test dataset consisted of the original images. The distribution of orientations in this split is shown in Fig. \ref{fig:original-orientation-test}.}
    \label{fig:app-mixed-predicted-orientations-first}
\end{figure}

\begin{figure}[H]
    \centering
    \subfloat[Normalized label count of 3.0]{
        \includegraphics[width=0.43\columnwidth]{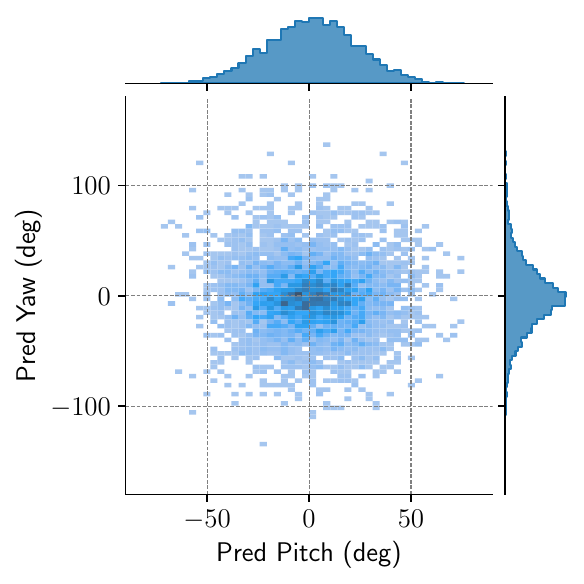}
        \label{fig:app-mixed-predicted-orientations-3000}
    }
    \hfill
    \subfloat[Normalized label count of 4.0]{
        \includegraphics[width=0.43\columnwidth]{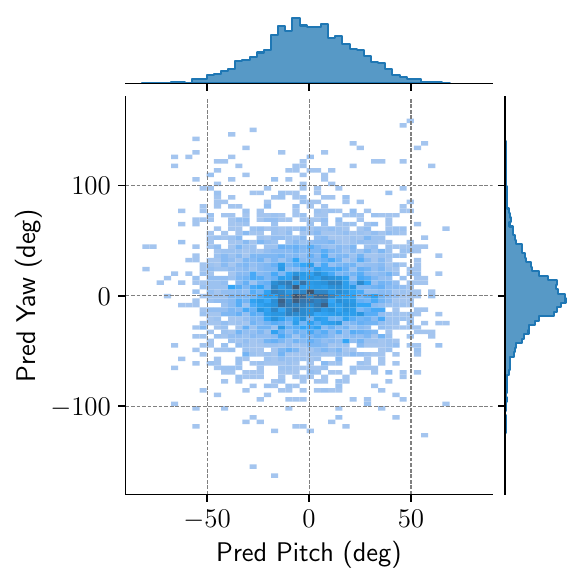}
        \label{fig:app-mixed-predicted-orientations-4000}
    }
    \hfill
    \subfloat[Normalized label count of 5.0]{
        \includegraphics[width=0.43\columnwidth]{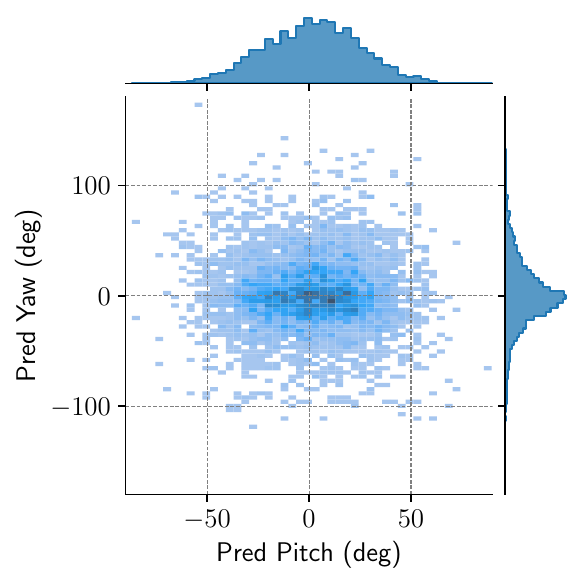}
        \label{fig:app-mixed-predicted-orientations-5000}
    }
    \caption{Figure showing the orientations predicted for different training dataset sizes, using both original and rendered images from a normalized label count of 3.0. The test dataset consisted of the original images. The distribution of orientations in this split is shown in Fig. \ref{fig:original-orientation-test}.}
    \label{fig:app-mixed-predicted-orientations-second}
\end{figure}

\end{document}